\documentclass[10pt,twocolumn,letterpaper]{article}

\usepackage{cvpr}
\usepackage{times}
\usepackage{epsfig}
\usepackage{graphicx}
\usepackage{amsmath}
\usepackage{amssymb}
\usepackage{xcolor}
\usepackage{makecell}
\usepackage{balance}
\usepackage{subcaption}

\cvprfinalcopy 

\def\cvprPaperID{15} 

\newcommand{\comment}[1]{}

\ifcvprfinal\fi
\begin{document}

\title{Compact Scene Graphs for Layout Composition and Patch Retrieval}
\author{Subarna Tripathi~~~Sharath Nittur Sridhar \thanks{Indicates Equal Contribution}~~~Sairam Sundaresan\footnotemark[1]~~~Hanlin Tang\\
Intel AI Lab,\\
San Diego, USA\\
{\tt\small \{subarna.tripathi, sharath.nittur.sridhar, sairam.sundaresan, hanlin.tang\}@intel.com}
}

\maketitle

\label{abs}
\begin{abstract}

Structured representations such as scene graphs serve as an efficient and compact representation that can be used for downstream rendering or retrieval tasks. However, existing efforts to generate realistic images from scene graphs perform poorly on scene composition for cluttered or complex scenes. 
We propose 
two 
contributions to improve the scene composition. 
First, we enhance the scene graph representation with heuristic-based relations, which add minimal storage overhead. 
Second, we use 
extreme points representation to supervise the learning of the scene composition network. 
These methods achieve significantly higher performance over existing work (69.0\% vs 51.2\% in relation score metric). 
We additionally demonstrate how scene graphs can be used to retrieve pose-constrained image patches that are semantically similar to the source query. Improving structured scene graph representations for rendering or retrieval is an important step towards realistic image generation.


\end{abstract}

\section{Introduction}
\label{sec:intro}
When we just need to communicate the semantic 'gist' of an image, and not necessarily the pixels, structured representations such as scene graphs are a compact alternative. Applications include image synthesis or scene composition from scene graphs \cite{img_gen_from_scene_graph18} and graph-based image patch retrieval. 

 For image synthesis, previous work has used representations such as class labels\cite{cgan}, captions \cite{ma_reed16}, or latent dimensions \cite{prog-gan}. However, these methods still struggle to generate realistic images across a broad vocabulary or for complex scene compositions.  Natural language representations such as captions require overcoming their inherently linear ordering to infer relationships. Scene graphs help alleviate this limitation of captions by providing a structured description of complex scenes. They can compactly model attributes, spatial relationships and hierarchy of various objects in a scene. However, using scene graphs for image generation has its own challenges as evidenced by Johnson \etal \cite{img_gen_from_scene_graph18}. While their approach yielded significantly improved images, it notably struggled with cluttered scenes or small objects. 
 
 Models such as Johnson \etal have two stages. The first stage uses the scene graph to generate a realistic scene layout, encoded as segmentation mask. The second stage uses the segmentation mask to synthesize realistic images. The second stage is well covered by existing work on image synthesis \cite{crn, SIMS_QiCJK18, park2019SPADE, Hong2018CaptionLayout, Gan_2018_CVPR}. The key point of failure in existing methods is the first stage -- generating realistic scene compositions.
 
To improve the ability to use compact scene graph representation for these tasks, in this paper we propose several improvements. We first improve the scene graph representation with heuristic-based relations. We employ a graph convolutional neural network to generate a scene layout that is compliant to the provided scene graph. Instead of regressing bounding boxes for each object in the scene, we regress the extreme points, which provide a tighter spatial bound compared to bounding boxes. By leveraging sparse shape information, these lightweight features allow us to efficiently retrieve the best matching candidate patches from an image database such as \cite{coco-stuff}. 

\begin{figure}[!t] \begin{center}
    \includegraphics[width=0.49\textwidth]{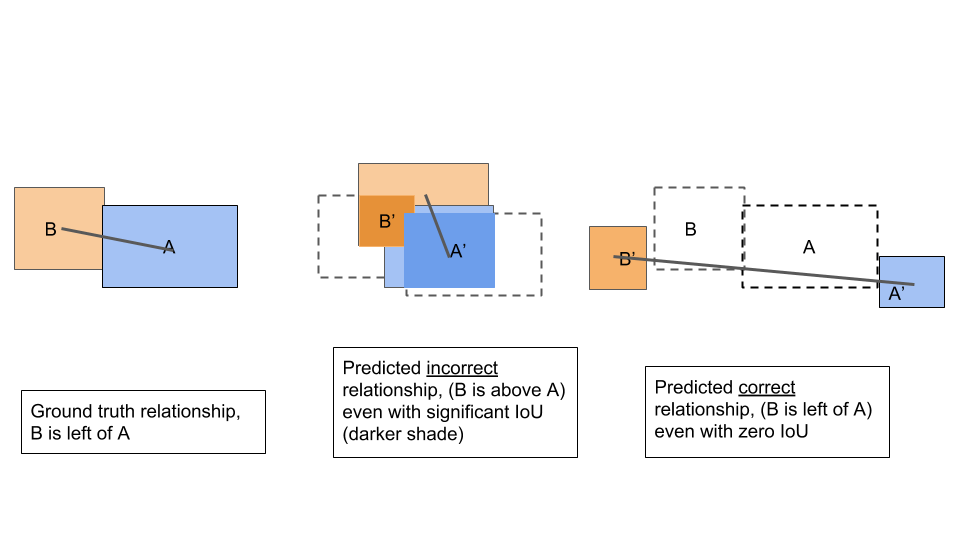}
\caption{Relation Score vs IoU. \emph{Left}: Example
ground truth relationship between two objects \textbf{A} and \textbf{B}. \emph{Center}: A predicted relationship between the objects \textbf{A} and \textbf{B} in which \textbf{B} is above \textbf{A}. Note that the high IoU here does not guarantee compliance with the intended geometric relationship. Intersection with ground truth is highlighted with darker shades. \emph{Right}: A predicted relationship between the objects \textbf{A} and \textbf{B} in which the two objects have no overlap. Note that despite the IoU being 0 here, the predicted relationship could still be compliant with the ground truth relationship. Both these scenarios would be accurately captured by the Relation Score but not the IoU.}  
\label{fig:rel_score}
\end{center}\end{figure}

Further, since this is a relatively new task, we leverage an evaluation metric suited to evaluate our scene composition method: the relation score \cite{scene_context_image_gen} which measures the compliance of the generated layout to the scene graph 
as shown in Figure \ref{fig:rel_score}.
Using the COCO-stuff \cite{coco-stuff} dataset, we evaluate our proposed model on composing scene layouts and on patch retrieval. Our experiments show that our proposed method establishes a new state of the art in scene layout generation from scene graphs, outperforming Johnson \etal \cite{img_gen_from_scene_graph18} qualitatively and quantitatively in layout prediction. Further, we are able to show the effectiveness of using extreme points to retrieve object patches which respect the scene graph. These improvements allow models to better leverage compact structured representations such as scene graphs. We leave the generation of images based on the predicted layout and retrieved patches for future work.

\section{Related Work}
\label{sec:related}
Scene graphs provide a compact and structured description of complex scenes \cite{battaglia2018relational} and the semantic relationship between objects. Ways to leverage this representation have generally fallen into three categories: generating graphs from images, generating images from graphs, and using graphs for image retrieval. Generating scene graphs from images is relatively well studied task, with approaches ranging from augmenting RNNs \cite{xu2017scenegraph} to re-purposing keypoint models \cite{hourglass_NewellYD16} to embedding-based approachs \cite{pixel_to_graph17}. 

Scene graphs have also been explored for image retrieval tasks \cite{jnt_embedding17,SG_for_IR_15}. 
These methods aim to search for an entire image corresponding to the input scene graph and can not be used for retrieving patches to synthesize new composites.  
Image based patch retrieval methods have also been studied in \cite{lalonde-siggraph-07} and \cite{TanBCOB17}. Of note here is the work of Zhao \etal \cite{composite_aware_search2018} who use a background image, class label and bounding box to query a database and retrieve the best matching patch to blend with the background image. While this method shows promising results, it needs the help of multiple input sources to find an appropriate match. Further, it is unclear if the method would scale efficiently when several objects are to be retrieved simultaneously. 

Image generation from scene graphs is relatively new. Johnson \etal \cite{img_gen_from_scene_graph18} extract objects and features from a scene graph with a graph convolutional neural network. A network then applies these features to predict a scene layout of objects, which are then used by a cascaded refinement network \cite{crn} to generate realistic images. While this is a novel end-to-end approach, it suffers from two major limitations. First, it performs poorly when the scene is very cluttered or has small objects. Second, the generating scene layouts are often non-sensical due to the sparsity in the scene graph representation. 

Our approach improves on previous methods by applying heuristic data augmentation to add missing relations to the scene graph representation, and by using extreme point regression to improve the layout generation.
Learning the extreme points 
allows us to generate tighter spatial bound for each object as compared to the bounding box \cite{img_gen_from_scene_graph18}.    
While previous approaches have focused on using scene graphs to retreive images, here we focus on retrieving individual patches that are still compliant with the rest of the scene graph. In combination, this provides a solid foundation for compositing and blending methods to generate a high resolution realistic image from the retrieved patches.


\section{Method}
\label{sec:method}
\begin{figure*}[!t] \begin{center}
\begin{subfigure}[b]{0.9\textwidth}
    \includegraphics[width=\textwidth]{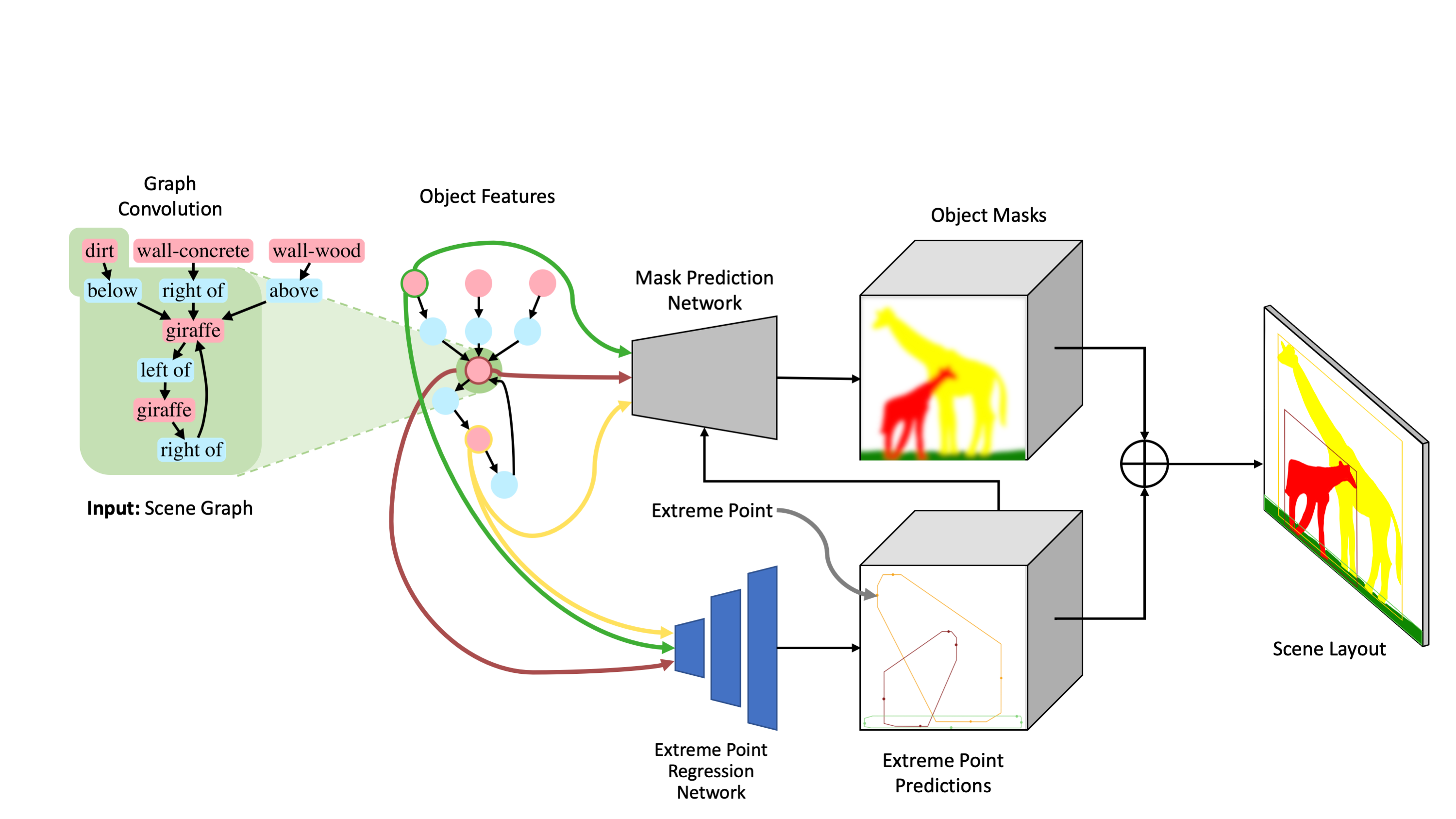}
  \end{subfigure}
  %
\caption{Overview of the proposed system - The input scene graph is passed through a graph convolutional neural network (GCNN) to process object class and relationship information present in nodes and edges respectively. The object embeddings thus generated are fed both to a mask prediction network as well as an extreme point regression network. The mask prediction network is conditioned on the output of the extreme point regression network. The octagonal representation of the extreme points and the generated object masks are then combined to produce the scene layout and bounding boxes. The extreme points thus produced are also used in the object patch retrieval process.}  
\label{fig:pipeline}
\end{center}\end{figure*}

The overall pipeline of the layout generation framework is illustrated in figure \ref{fig:pipeline}. Given a scene graph consisting of objects and their relationships, our model constructs a realistic layout corresponding to the scene graph. Our framework is inspired by \cite{img_gen_from_scene_graph18}. Briefly, the scene graph is converted into $D$-dimensional object embedding vectors ($o$) from a Graph Convolution Neural Network (GCNN), which are then passed to a mask prediction network as well as an extreme point regression network. The extreme points generated thus are used to condition the mask prediction. 
For each object $i$, 
we first achieve bounding box $r_i$ from its $4$ predicted extreme points. 
We then expand the object embedding vectors $o_i \in R^D$ to shape
$D \times 16 \times 16$, and warp it to the position of the bounding
box $r_i$ using bilinear interpolation to give an object layout
${o_i}^
{layout}
 \in R^{D\times H \times W} $ , where $H = 256$ and $W = 256$ are the
output spatial layout dimensions. We sum all object layouts to obtain
the scene layout 
$S^{layout} =\sum_{i}{o_i}^{layout}$
.
Finally, using the computed extreme points for each object, we retrieve individual patches which best respects the segmentation mask for that object. Each of these steps are described below in more detail.

\paragraph{Heuristic-based Data Augmentation} The scene graph representation, while compact, is often incomplete, leading to poor scene composition layouts. We used heuristics to augment the scene graph representation with new spatial relations that induce a richer learned representation as shown in Figure \ref{fig:rel_augmentation}. We quasi-exhaustively determined the depth order between objects from observers' viewpoint. For 2D images, determining this order is non-trivial. We utilize linear perspective based heuristics instead for augmenting spatial relationship vocabulary. We provide the details in the experiments section.   

\newcommand{\maxspace}{\textwidth}
\begin{figure}[hbt!]
\captionsetup[subfigure]{justification=centering}
\centering 
  \begin{subfigure}{0.2\maxspace}
    \includegraphics[height=20mm,scale=1]{images/augmentation/instance-instance/000116_gt_img.png}
  \end{subfigure}
  \begin{subfigure}{0.2\maxspace}
    \includegraphics[height=12mm,scale=0.5]{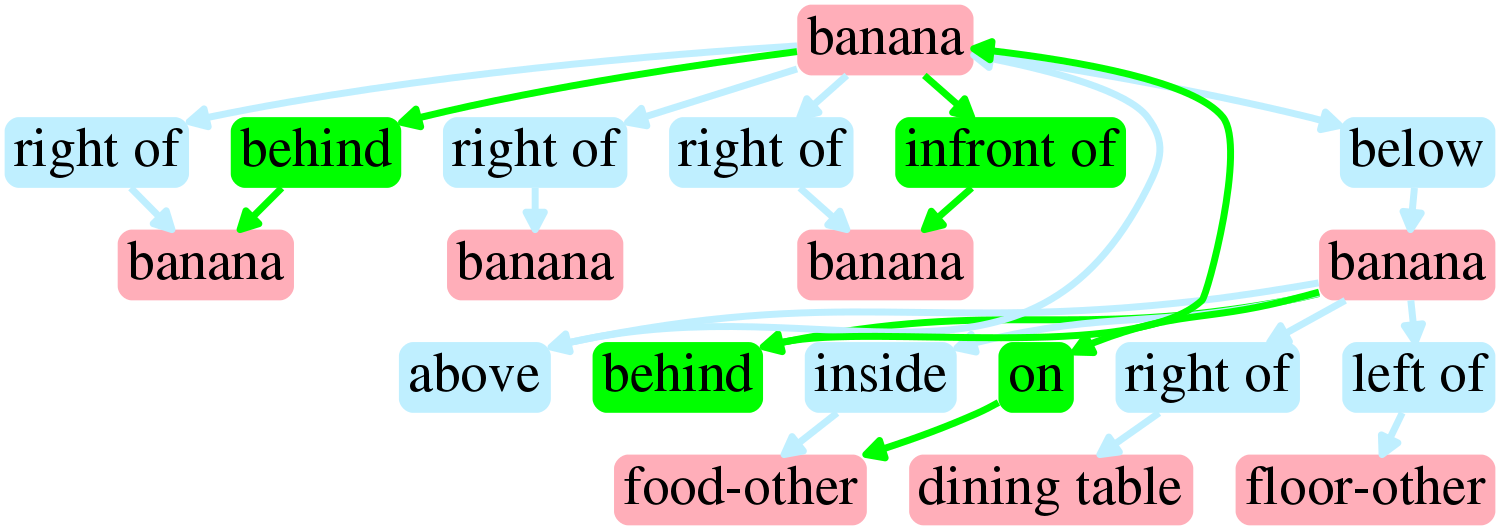}
  \end{subfigure}
  
  \begin{subfigure}{0.2\maxspace}
    \includegraphics[height=20mm,scale=1]{images/augmentation/instance-stuff/000187_gt_img.png}
  \end{subfigure}
  \begin{subfigure}{0.2\maxspace}
    \includegraphics[height=15mm,scale=0.5]{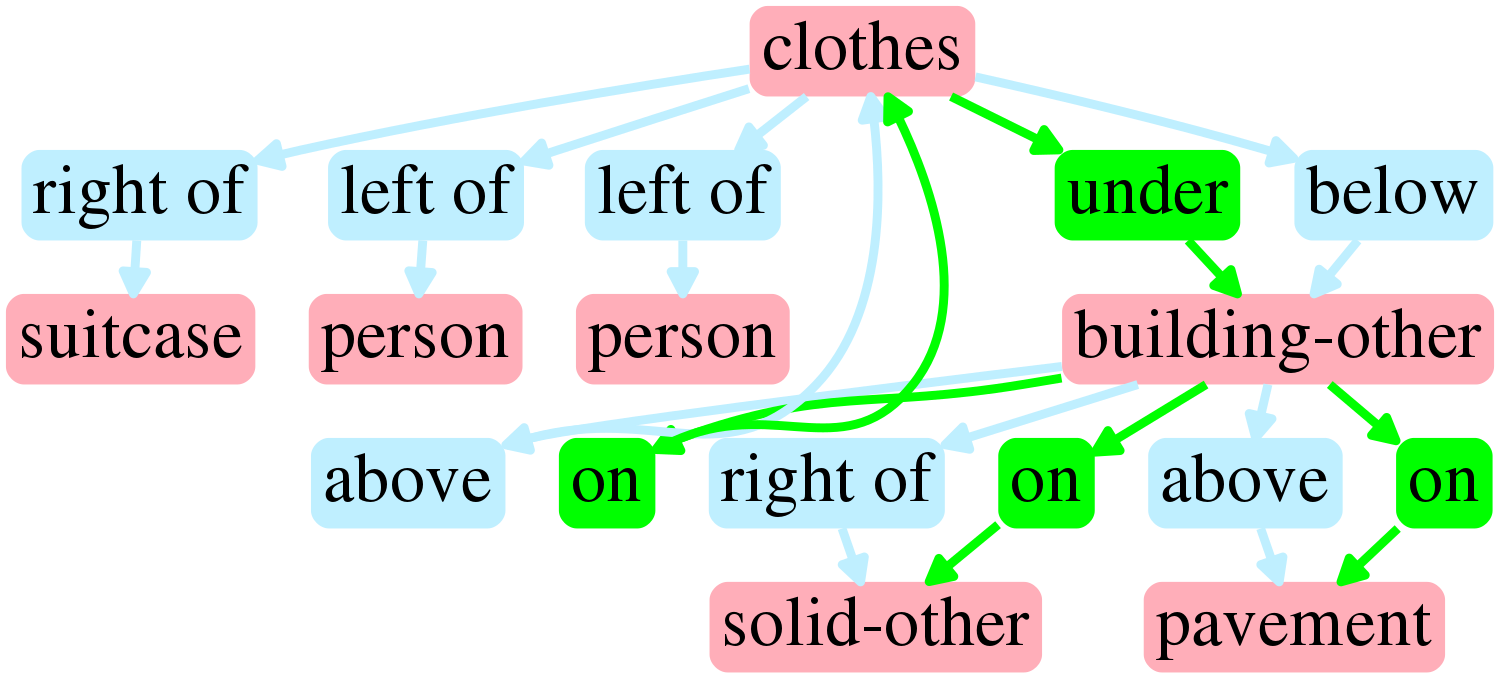}
  \end{subfigure}
  
  \begin{subfigure}{0.2\maxspace}
    \includegraphics[height=20mm,scale=1]{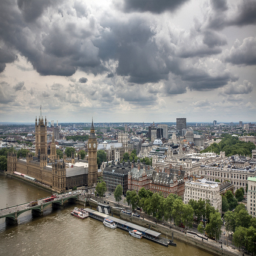}
  \end{subfigure}
  \begin{subfigure}{0.2\maxspace}
    \includegraphics[height=15mm,scale=0.5]{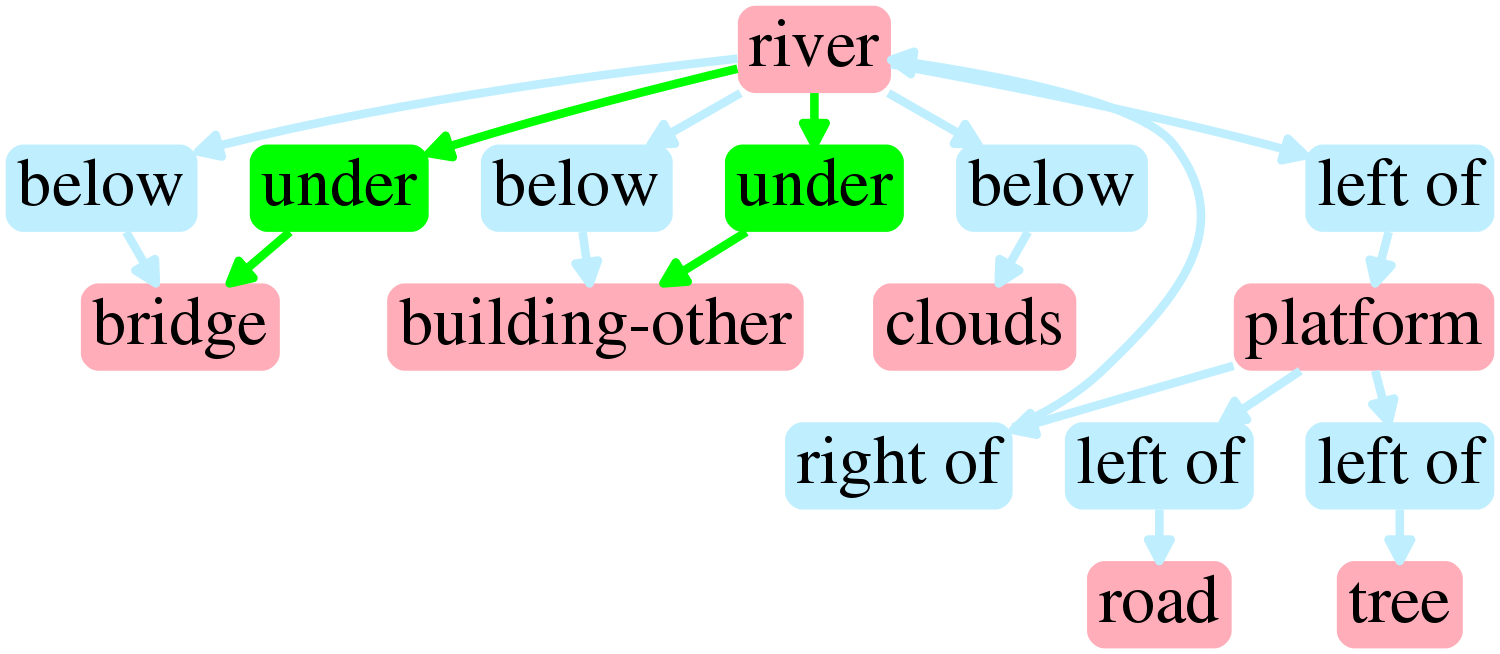}
  \end{subfigure}
  
\caption{Relationship augmentation based on heuristics for (a) thing-thing relationship consisting of `in front of' and `behind' based on perspective geometry heuristic, (b) thing-stuff relationship including `under' and `on' and (c) stuff-stuff relationship. Augmented relationships are in `green', while regular spatial relationships are in `light blue'. Best viewed in color.}  
\label{fig:rel_augmentation}
\end{figure}

\paragraph{Extreme point supervision for representation learning}
Extreme points proposed by Zhou \etal \cite{extremenet2019} have shown to be successful in a bottom-up object detection framework. They provide a compact way of representing the shape of an object. Per \cite{extremenet2019}, given the bounding box $(x^{(l)}, y^{(t)}, x^{(r)}, y^{(b)})$ of an object, an extreme point is a point such that no other point $(x^{(i)},y^{(i)})$ on the object lies further along one of the four cardinal directions: top, bottom, left, right (For example, $(x^{(t)}, y^{(t)})$, $(x^{(b)}, y^{(b)})$, $(x^{(l)}, y^{(l)})$, $(x^{(r)}, y^{(r)})$ would be the four extreme points for the bounding box defined above). Extreme point prediction from scene graphs however, is more challenging since there are no visual features to leverage. We use an extreme point regression network which predicts extreme points with the aim of creating a layout that obeys the relationships described by the scene graph. Further, these points enable the construction of an octagon around each object in the layout which can then be used as a sparse shape representation for the object. Using this representation, and the object class, we can retrieve a set of matching patches which would best fit the segmentation mask for each object. 


\paragraph{Network architecture} 

Specifically, our network consists of three components, a graph convolutional network, a mask prediction network and an extreme point regression network.

\begin{itemize}
\item \textbf{\emph{Graph Convolutional Network}}: In order to produce a scene layout, the input scene graph must first be transformed from the graph domain into a format which can be used by the mask prediction and extreme point regression networks. A graph convolutional network facilitates this transformation by processing the input scene graph and generating an embedding vector as its output. Like \cite{img_gen_from_scene_graph18}, our graph convolutional network has 5 layers with an input dimension $D_{in}$ of 128, and an output dimension $D_{out}$ of 128 with a hidden dimension of 512.

\item \textbf{\emph{Extreme Point Regression Network}}: A key part of predicting a well defined layout is the ability to produce tight bounds on each object in the scene. For this, we use an extreme point regression network which consists of two fully connected layers with ReLU activation after the first layer. The network takes as input the final embedding vector produced for each object and instead of regressing 4 points corresponding to the bounding box of the object, regresses 10 points corresponding to the 4 extreme points of the object and the center of the object. These points are then used to condition the mask prediction. 

\item \textbf{\emph{Mask Prediction Network}}: For each object, the mask prediction network predicts a segmentation mask.
It takes as input the final embedding vector for each object along with the extreme points which have been regressed for that object
concatenated as a vector. 
The mask prediction network consists of a sequence of four sub-blocks, 
where each sub-block has a $2\times 2$ nearest neighbour upsampling step followed by batch normalization and a $3\times 3$ convolution layer with ReLU activation. Similar to \cite{img_gen_from_scene_graph18}, the network ends with a $1\times 1$ convolution layer and sigmoid activation. The output 
is a $K\times K$ mask with each value of the mask in the range of $[0,1]$.\\ 
\end{itemize}

\paragraph{Training}
The layout generation network is trained to minimize the weighted sum of two losses:
\begin{itemize}
    \item \textbf{\emph{Extreme point loss}}, $\mathcal{L}_{ep}$ penalizes the $\ell_\emph{2}$ difference between the ground truth and predicted extreme point co-ordinates as $\mathcal{L}_{ep}=\sum_{i=1}^n\|X_i - \hat X_i\|_2$ where $\hat X_i$ is a predicted extreme point and $X_i$ is the corresponding ground truth extreme point.
    
    \item \textbf{\emph{Mask loss}}, $\mathcal{L}_{mask}$, penalizes differences between the ground truth and predicted masks with pixel-wise cross entropy.

    The total loss is then computed as a weighted sum of the two losses above
    
    \begin{equation}
        \mathcal{L}_{total} = \lambda_{1}\mathcal{L}_{ep} + \lambda_{2}\mathcal{L}_{mask}
        \label{eq:total_loss}
    \end{equation}
    
    Here, $\lambda_{1}$ and $\lambda_{2}$ are weights applied to each loss term.
    
    
    
    
    
    
    
\end{itemize}

In our experiments, we found that using the Adam optimizer \cite{adam_KingmaB14} with a learning rate of $0.0001$ and $\beta_1$ of $0.9$ for a batch size of $32$ gave us optimal results. Further, we set $\lambda_{1}$ to $10$ and $\lambda_{2}$ to $0.1$. 

\paragraph{Shape-aware retrieval}
In our patch retrieval task, the algorithm is tasked to return patches similar to the ground truth objects in the query scene graphs. Importantly, the retrieval algorithm only has access to the query scene graph, and not the associated ground truth image. With only the compact representation of the scene graph, it must use the scene graph context for the query. Our algorithm first predicts the extreme points of the object, then uses a $L_2$ metric on the extreme point octagon to retrieve similar patches in the training set. We measured performance as Intersection-over-Union (IoU) between the ground truth patch object mask and the retrieved patch object mask. 

We benchmarked several retrieval algorithms: (1) bounding-box based $\ell_\emph{2}$ retrieval, (2) our extreme points based $\ell_\emph{2}$ retrieval, and (3) random retrieval.

\section{Experiments}
\label{sec:results}
We train our model on the COCO-Stuff\cite{coco-stuff} dataset to generate $256\times 256$ scene layouts . 
In our experiments, we aim to show that the generated layouts look realistic and that they respect the objects and relationships of the input scene graph. We further show that our predicted extreme points can be used for shape aware object patch retrieval across $23$ common object categories. The following subsections elaborate on our dataset, as well as qualitative and quantitative results.

\begin{table}[]
\small
\centering
\caption{Relationship compliance. Relation Score (Higher is better) on the COCO-Stuff dataset. DA denotes data augmentation. 
EP stands for extreme points based supervision. 
}
\begin{center}\begin{tabular}{ccccc} 
\textbf{Model} & \textbf{Relation Score} &    \textbf{Avg IOU} \\
\hline
Johnson \etal \cite{img_gen_from_scene_graph18} & 51.2\% & 45.9\% \\
Johnson \etal \cite{img_gen_from_scene_graph18}+ DA & 59.8\% & 51.0\% \\
EP +$\ell_\emph{2}$ (Ours) & \textbf{69.0\%} & \textbf{51.6\%} \\
\hline
\end{tabular}
\end{center}
\label{tab:rel_score} 
\end{table}

\subsection{Datasets}

\textbf{COCO}: We performed experiments on the 2017 COCO-Stuff
dataset \cite{coco-stuff}, which augments a subset of the COCO
dataset \cite{coco} with additional stuff categories. The dataset annotates
$40K$ train and $5K$ val images with bounding boxes
and segmentation masks for $80$ thing categories (people,
cars, etc.) and $91$ stuff categories (sky, grass, etc.).
Similar to \cite{img_gen_from_scene_graph18},  we used thing and stuff annotations to construct synthetic scene
graphs based on the 2D image coordinates of the objects,
encoding six mutually exclusive geometric relationships: `left'
`of', `right of', `above', `below', `inside', and `surrounding'.
We ignored objects covering less than 2\% of the image,
and used images with $3$ to $8$ objects.

In the augmentation process, we exploit the `thing' (`object`) and `stuff' annotations. 
We encode two heuristics based relationships: `behind' and `in front of' between two spatially overlapping `things'. 
Among overlapping `things' A and B, A is `in front of' B if the bottom boundary of A's bounding box is closer to the image's bottom edge.  
Additionally, `on' and `under' relationships between overlapping `thing' and `stuff' are imposed. A `thing' is always `on' the `stuff' from viewer's perspective. Figure \ref{fig:rel_augmentation} shows examples of between instances and between instance-stuff relationship augmentation.


\subsection{Scene Layout Prediction Results}
For COCO, we compare the predicted layout with the ground truth using both Intersection-over-Union (IoU) and Relation Score (Figure \ref{fig:rel_score}). 
It is to be noted that relation score is more suited metric than IOU for this task. 
By applying extreme point based supervision, data augmentation and location conditioned mask prediction, we significantly improve the performance of the model (Table \ref{tab:rel_score}) for both metrics. The extreme points annotations are extracted from the segmentation masks from \emph{objects} classes. For \emph{stuff} classes, we created trivial extreme points from the bounding box annotations.

\begin{figure*}[hbt!]
    \captionsetup[subfigure]{justification=centering}
	\centering 
	\begin{subfigure}{.20\textwidth}
		\centering
		\includegraphics[height=20mm,scale=0.2]{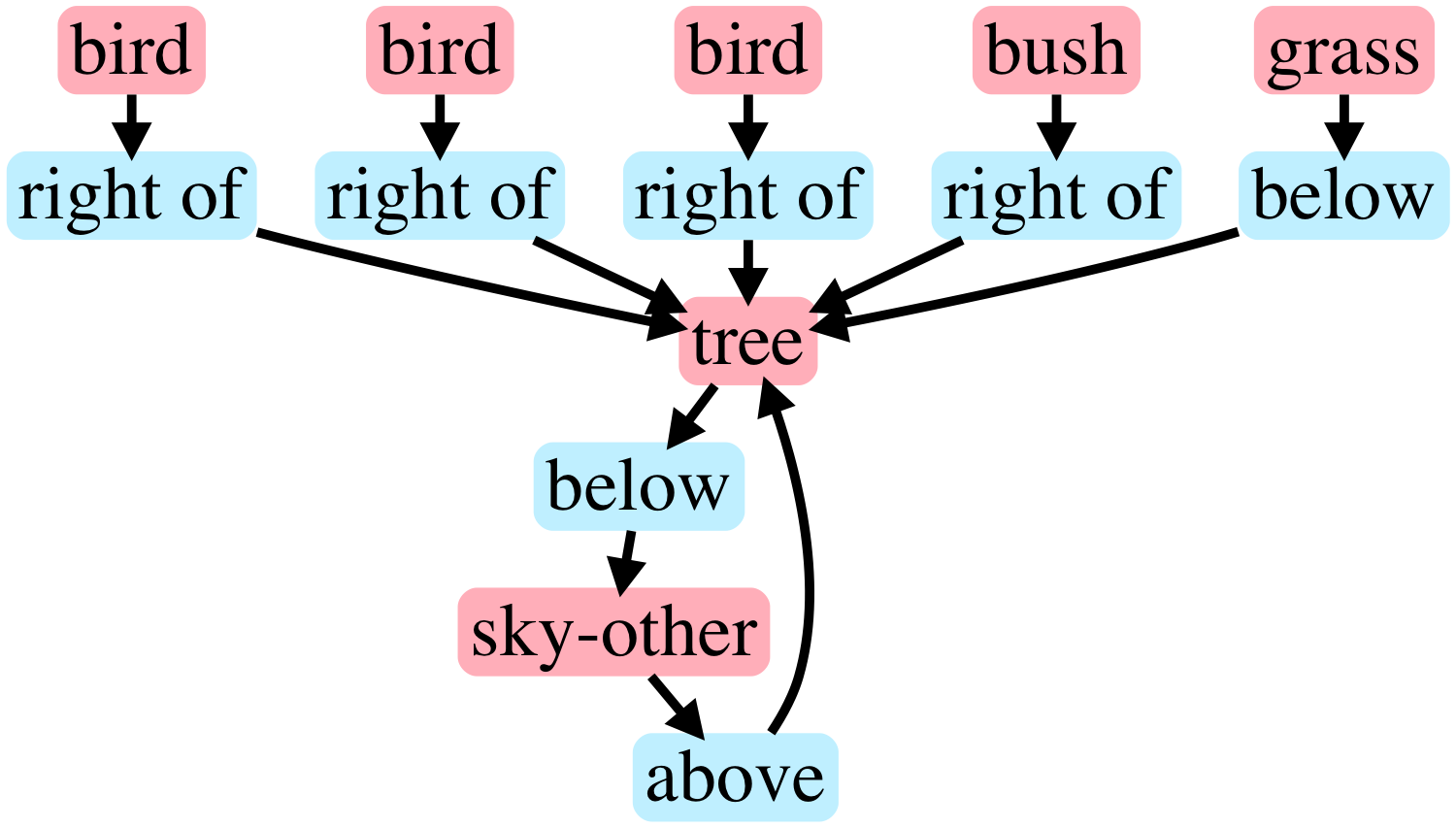}
	\end{subfigure}%
	\begin{subfigure}{.16\textwidth}
	    \centering
		\includegraphics[height=20mm,scale=1]{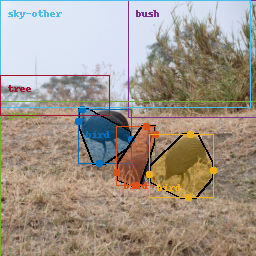}
	\end{subfigure}%
	\begin{subfigure}{.16\textwidth}
	    \centering 
		\includegraphics[width=20mm,scale=1]{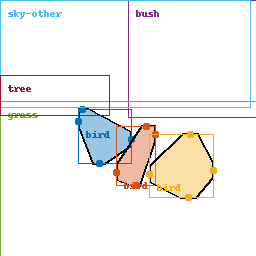}
	\end{subfigure}%
	\begin{subfigure}{.16\textwidth}
	    \centering
		\includegraphics[height=20mm,scale=1]{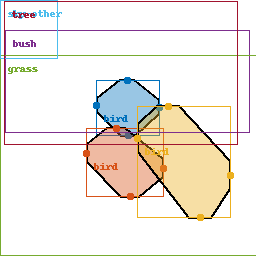}
	\end{subfigure}%
	\begin{subfigure}{.16\textwidth}
	    \centering
		\includegraphics[height=20mm,scale=1]{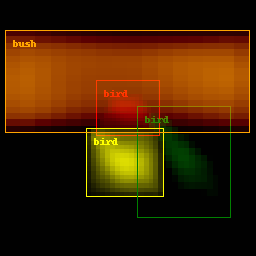}
	\end{subfigure}%
	\begin{subfigure}{.16\textwidth}
	    \centering
		\includegraphics[height=20mm,scale=1]{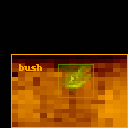}
	\end{subfigure}%
	\par\bigskip
	\begin{subfigure}{.20\textwidth}
		\centering
		\includegraphics[height=20mm,scale=0.2]{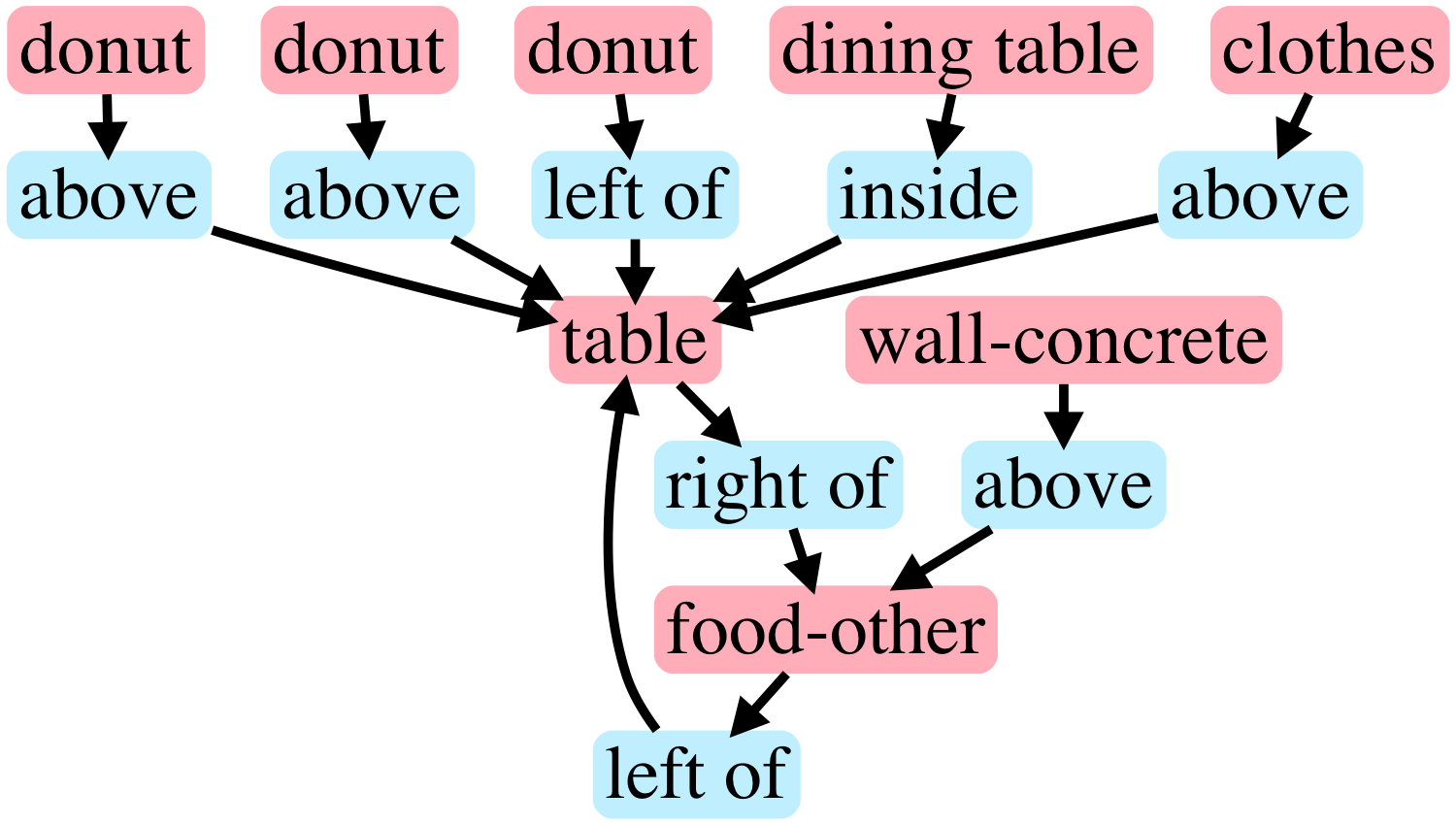}
	\end{subfigure}%
	\begin{subfigure}{.16\textwidth}
	    \centering
		\includegraphics[height=20mm,scale=1]{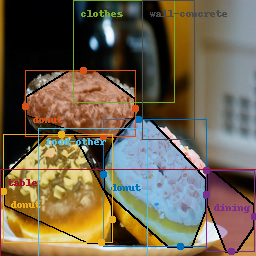}
	\end{subfigure}%
	\begin{subfigure}{.16\textwidth}
	    \centering 
		\includegraphics[width=20mm,scale=1]{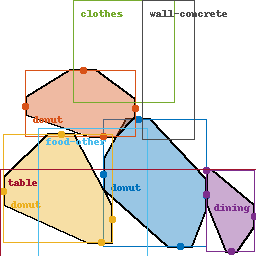}
	\end{subfigure}%
	\begin{subfigure}{.16\textwidth}
	    \centering
		\includegraphics[height=20mm,scale=1]{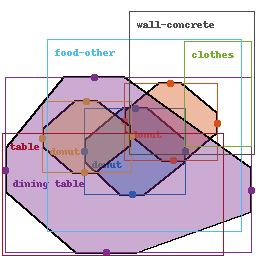}
	\end{subfigure}%
	\begin{subfigure}{.16\textwidth}
	    \centering
		\includegraphics[height=20mm,scale=1]{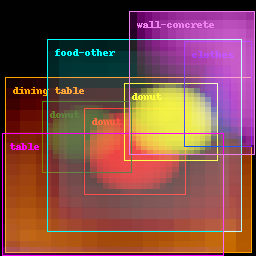}
	\end{subfigure}%
	\begin{subfigure}{.16\textwidth}
	    \centering
		\includegraphics[height=20mm,scale=1]{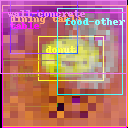}
	\end{subfigure}%
	\par\bigskip
	\begin{subfigure}{.20\textwidth}
		\centering
		\includegraphics[height=20mm,scale=0.2]{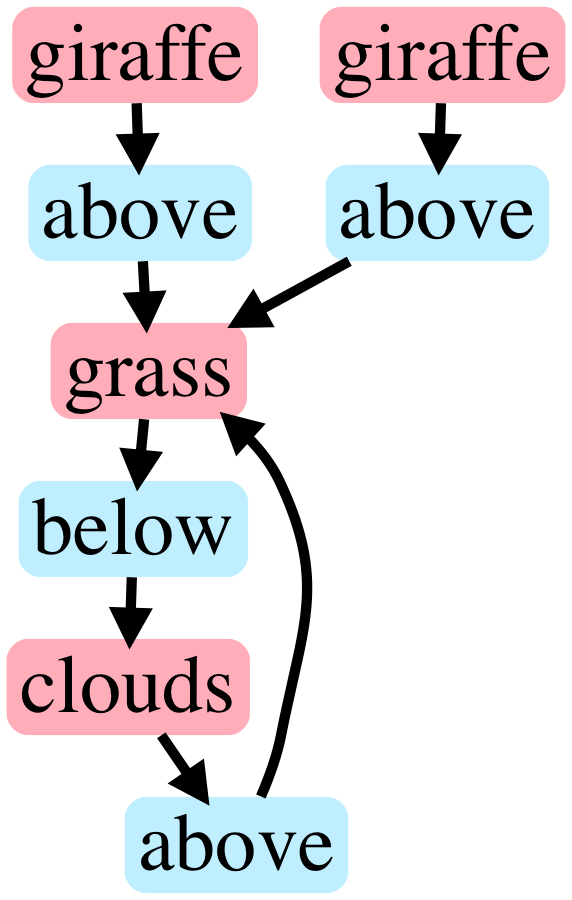}
	\end{subfigure}%
	\begin{subfigure}{.16\textwidth}
	    \centering
		\includegraphics[height=20mm,scale=1]{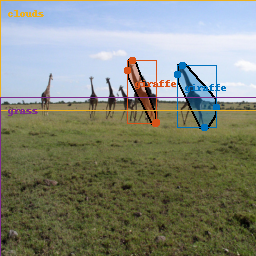}
	\end{subfigure}%
	\begin{subfigure}{.16\textwidth}
	    \centering 
		\includegraphics[width=20mm,scale=1]{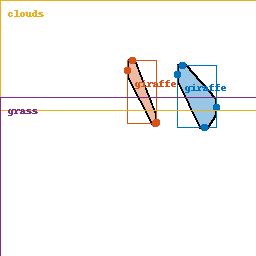}
	\end{subfigure}%
	\begin{subfigure}{.16\textwidth}
	    \centering
		\includegraphics[height=20mm,scale=1]{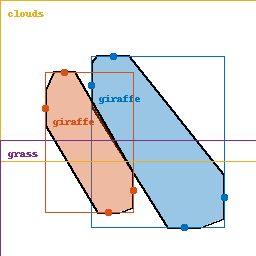}
	\end{subfigure}%
	\begin{subfigure}{.16\textwidth}
	    \centering
		\includegraphics[height=20mm,scale=1]{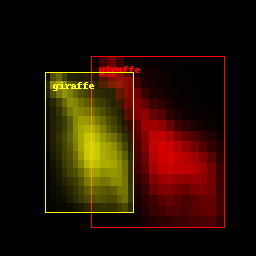}
	\end{subfigure}%
	\begin{subfigure}{.16\textwidth}
	    \centering
		\includegraphics[height=20mm,scale=1]{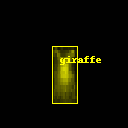}
	\end{subfigure}%
	\par\bigskip
	\begin{subfigure}{.20\textwidth}
		\centering
		\includegraphics[height=20mm,scale=0.2]{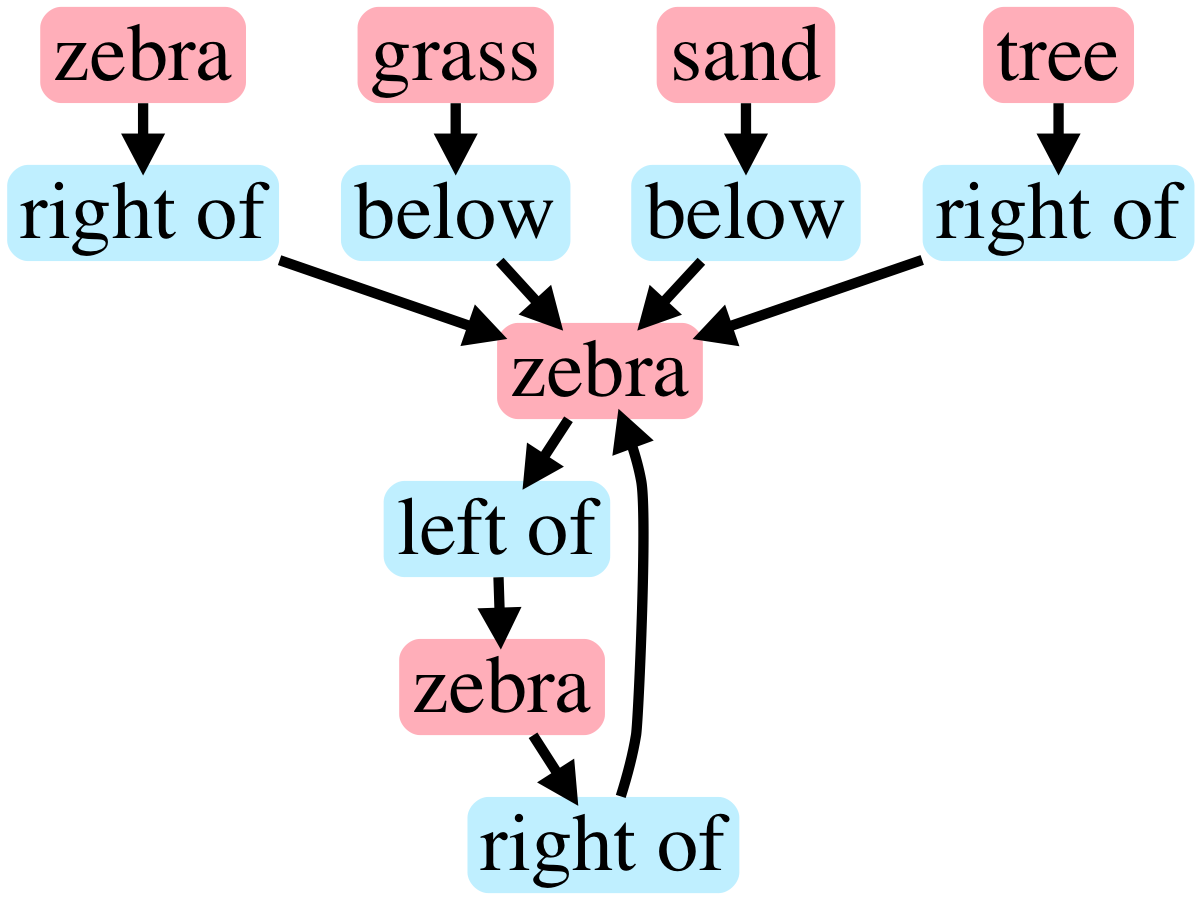}
	\end{subfigure}%
	\begin{subfigure}{.16\textwidth}
	    \centering
		\includegraphics[height=20mm,scale=1]{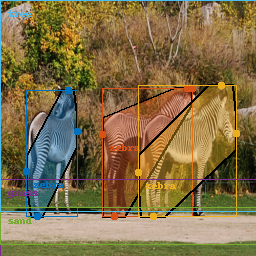}
	\end{subfigure}%
	\begin{subfigure}{.16\textwidth}
	    \centering 
		\includegraphics[width=20mm,scale=1]{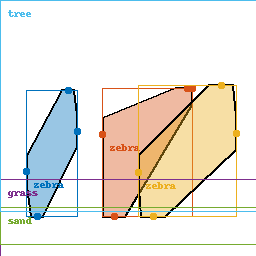}
	\end{subfigure}%
	\begin{subfigure}{.16\textwidth}
	    \centering
		\includegraphics[height=20mm,scale=1]{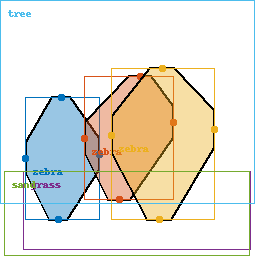}
	\end{subfigure}%
	\begin{subfigure}{.16\textwidth}
	    \centering
		\includegraphics[height=20mm,scale=1]{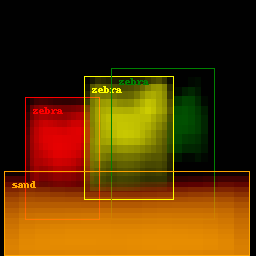}
	\end{subfigure}%
	\begin{subfigure}{.16\textwidth}
	    \centering
		\includegraphics[height=20mm,scale=1]{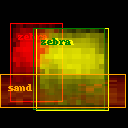}
	\end{subfigure}%
	\par\bigskip
	\begin{subfigure}{.20\textwidth}
		\centering
		\includegraphics[height=20mm,scale=0.2]{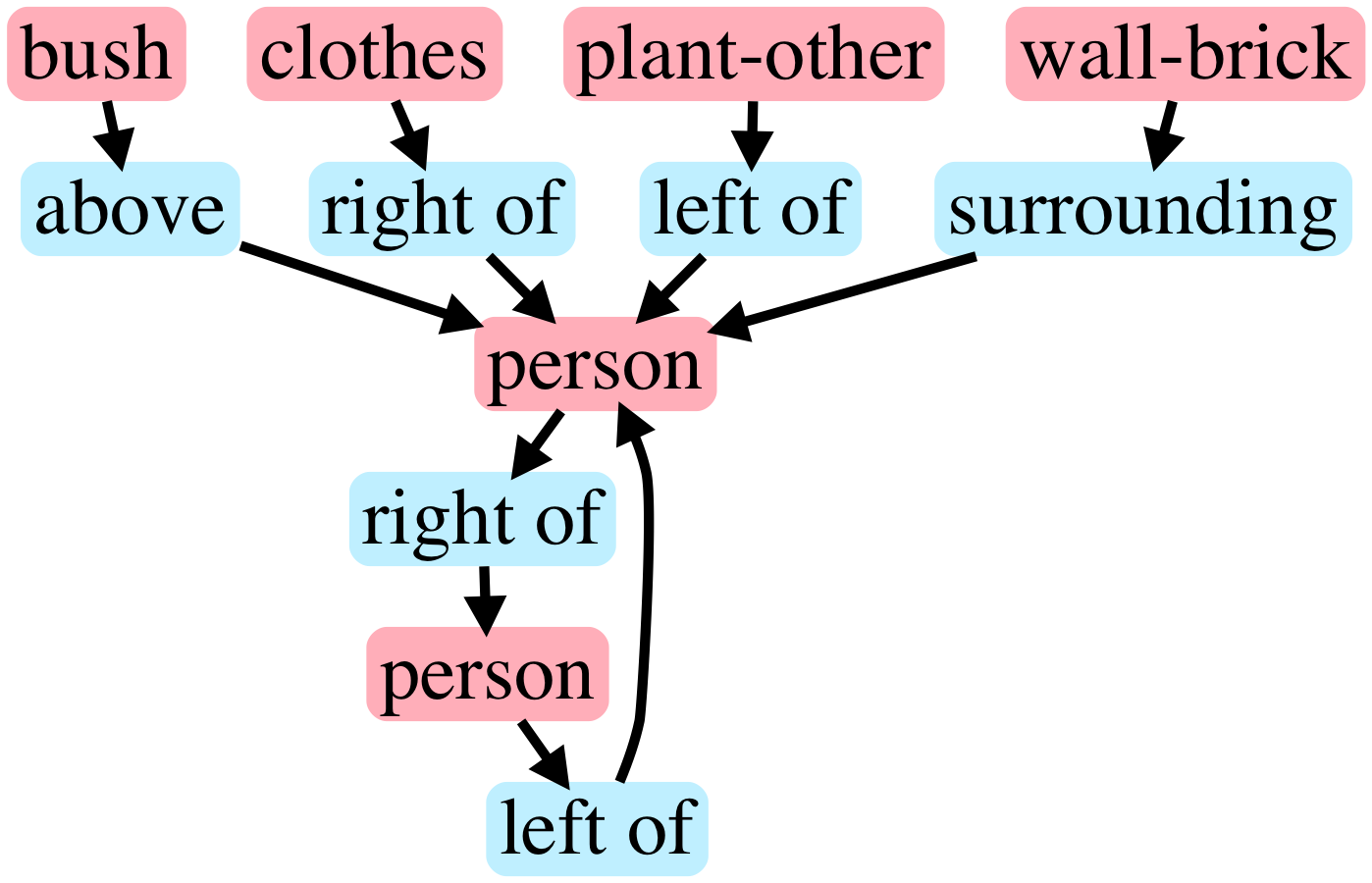}
	\end{subfigure}%
	\begin{subfigure}{.16\textwidth}
	    \centering
		\includegraphics[height=20mm,scale=1]{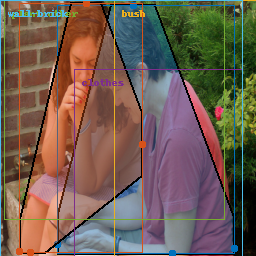}
	\end{subfigure}%
	\begin{subfigure}{.16\textwidth}
	    \centering 
		\includegraphics[width=20mm,scale=1]{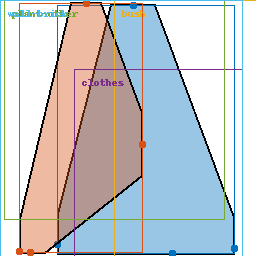}
	\end{subfigure}%
	\begin{subfigure}{.16\textwidth}
	    \centering
		\includegraphics[height=20mm,scale=1]{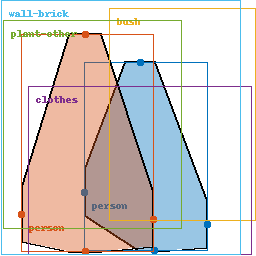}
	\end{subfigure}%
	\begin{subfigure}{.16\textwidth}
	    \centering
		\includegraphics[height=20mm,scale=1]{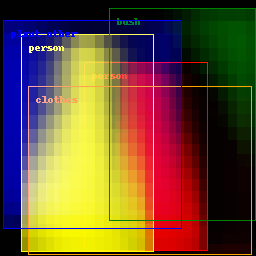}
	\end{subfigure}%
	\begin{subfigure}{.16\textwidth}
	    \centering
		\includegraphics[height=20mm,scale=1]{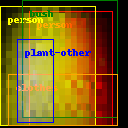}
	\end{subfigure}%
    
	\begin{subfigure}{.20\textwidth}
		\centering
		\includegraphics[height=18mm,scale=0.1]{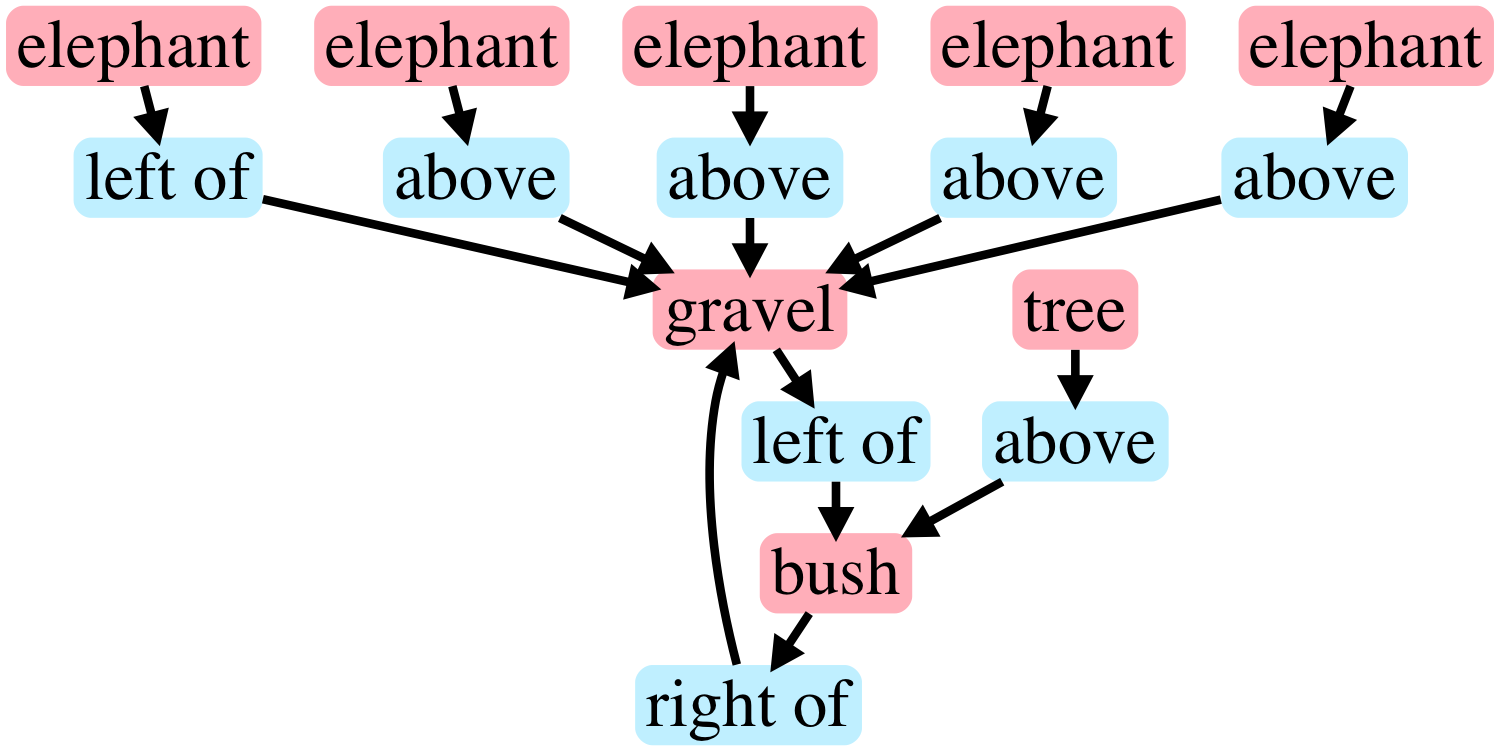}
		\par\medskip
		\caption{Scene Graph}
	\end{subfigure}%
	\begin{subfigure}{.16\textwidth}
	    \centering
	    \par\bigskip
		\includegraphics[height=20mm,scale=1]{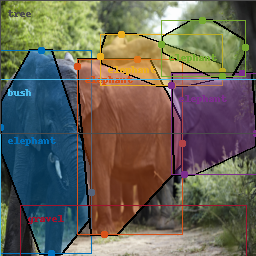}
		\caption{Ground Truth Image}
	\end{subfigure}%
	\begin{subfigure}{.16\textwidth}
	    \centering
	    \par\bigskip
		\includegraphics[width=20mm,scale=1]{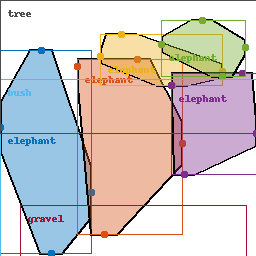}
		\centering\caption{Ground Truth Extreme Points}
	\end{subfigure}%
	\begin{subfigure}{.16\textwidth}
	    \centering
	    \par\bigskip
		\includegraphics[height=20mm,scale=1]{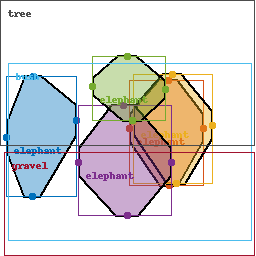}
		\caption{Predicted Extreme Points}
	\end{subfigure}%
	\begin{subfigure}{.16\textwidth}
	    \centering
		\includegraphics[height=20mm,scale=1]{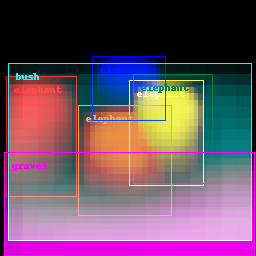} 
		\caption{\textbf{Our Result}} 
	\end{subfigure}%
	\begin{subfigure}{.16\textwidth}
	    \centering
		\includegraphics[height=20mm,scale=1]{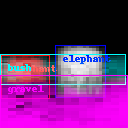} 
		\caption{Johnson \etal \cite{img_gen_from_scene_graph18}} 
	\end{subfigure}%
	\vspace{-0.75\baselineskip}
	\par\medskip
	\caption{Extreme point supervision and extreme point conditioned mask prediction yields improved layout prediction from scene graph. From left to right for each row - (a) Input scene graph, (b) Ground truth octagons from the COCO Stuff\cite{coco-stuff} dataset overlaid on the corresponding ground truth image, (c) Ground truth octagons shown separately, (d) Octagons generated from the predicted extreme points per object using the proposed method, (e) The scene layout mask predicted by our model v/s (f) The layout predicted by Johnson \etal ~\cite{img_gen_from_scene_graph18}. Note that the ground truth extreme points and their connecting octagons are overlaid on the corresponding ground truth image, are shown for reference. The proposed model handles the placement of multiple objects, inter-class spatial arrangement and learns better shapes overall.}
	\label{fig:scene_layout_results}
\end{figure*}

\begin{table}[]
\small
\caption{Retrieval Metric: Top-1 and Top-5 IoU scores averaged across 23 categories in the ground truth validation set. EP + $\ell_\emph{2}$ loss represents our proposed retrieval method which uses $\ell_\emph{2}$ loss on extreme points for matching. BB + $\ell_\emph{2}$ loss uses $\ell_\emph{2}$ loss on bounding boxes instead of extreme points and Random indicates a random selection of patches from the database for a given object category. } 
\begin{center}\begin{tabular}{ccc} 
\textbf{Retrieval Method} & \textbf{Top-1 IOU} &    \textbf{Top-5 IOU} \\
\hline
EP + $\ell_\emph{2}$  & \textbf{54\%} & \textbf{64\%} \\
BB + $\ell_\emph{2}$ & 44\% & 57\% \\
Random & 42\% & 55\% \\
\hline
\end{tabular}
\end{center}
\label{tab:retrieval}
\end{table}

Figure \ref{fig:scene_layout_results} shows visual results. Ground truth images with octagonal masks around the extreme points show the tighter bound on objects comparing with that of bounding box. We show the scene layout from our model and Johnson \etal model respectively. Our model places objects better especially if there are missing directed edges between objects in the input scene graph.   

\subsection{Retrieval Database Description}
We create a database of object patches extracted from the COCO  training dataset for a pre-defined set of $23$ classes using the bounding box co-ordinates. For each of the object patches, we also include a set of normalized extreme points \cite{extremenet2019} which will be used for matching and patch retrieval. The number of patches for each object varies between 1051 and 15292 images. 
The list of selected categories and the number of object patches in each category are presented in Table \ref{tab:ret_data}.

\begin{table*}[]
\centering
\caption{Retrieval Database: Shows the number of object patches in the database for different categories.}
\renewcommand{\arraystretch}{1.4}
\renewcommand{\tabcolsep}{1.4mm}
\scriptsize
\resizebox{\linewidth}{!}{
\begin{tabular}{@{}lr*{20}{c}c@{}}
\hline 
\textbf{bear} & \textbf{bed} & \textbf{bench} & \textbf{bicycle} & \textbf{bird} & \textbf{boat} & \textbf{bottle}  & \textbf{bus} & \textbf{car} & \textbf{cat} & \textbf{chair} & \textbf{cow} &  \textbf{dining table} &\textbf{dog} & \textbf{elephant} & \textbf{fire hydrant} & \textbf{giraffe} & \textbf{horse} & \textbf{motorcycle} & \textbf{stop sign} & \textbf{surfboard} & \textbf{train} & \textbf{zebra} \\
\hline
\hline
1152 & 4048 & 4382 & 3012 & 2966 & 3614  & 3831 & 4278 & 8600 & 4366 & 15292 & 3540 & 12201 & 4230
& 3947 & 1180 & 4454 & 4197 &  5057 & 1051 & 2676 & 4126 & 3909 \\
\hline
\end{tabular}
}
\label{tab:ret_data}
\end{table*}

\subsection{Experimental Results of Patch Retrieval}
\begin{figure*}[hbt!]
	\centering
	\captionsetup[subfigure]{justification=centering}
	\begin{subfigure}{.2\maxspace}
	    \centering
		\includegraphics[height=20mm,scale=0.2]{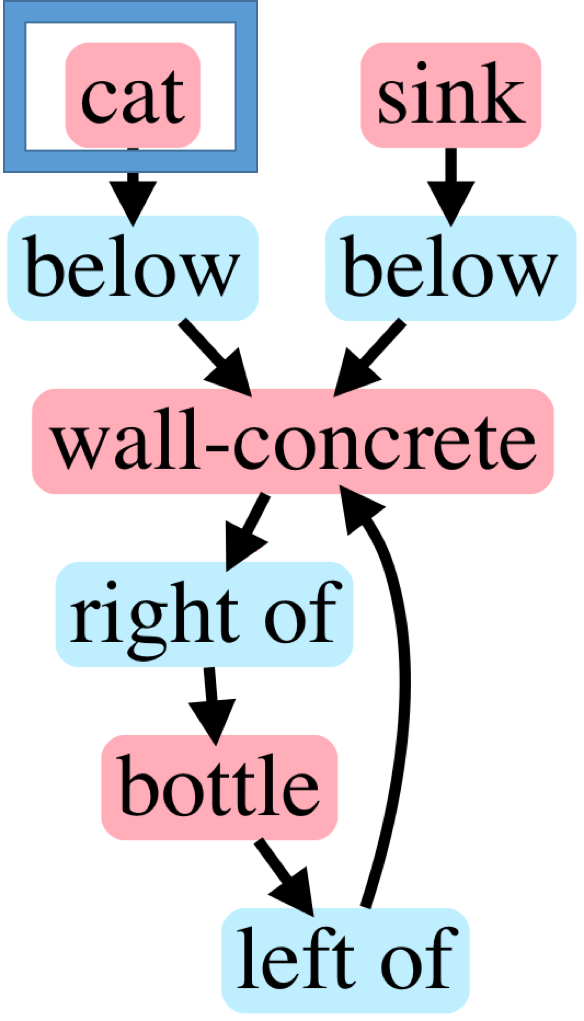}
	\end{subfigure}%
	\begin{subfigure}{.16\maxspace}
		\centering
		\includegraphics[height=20mm,scale=0.3]{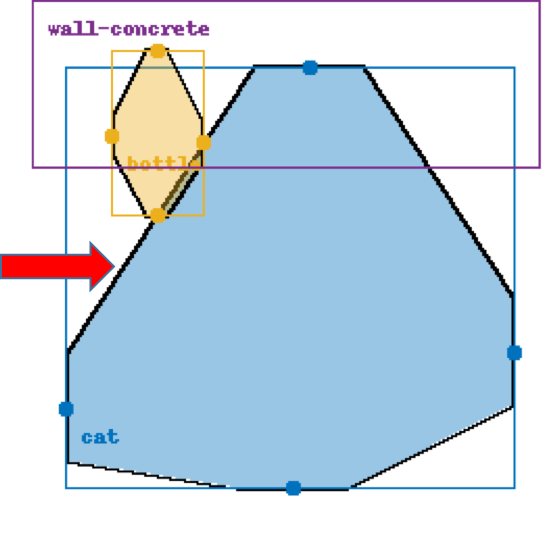}
	\end{subfigure}%
	\begin{subfigure}{.16\maxspace}
	    \centering
		\includegraphics[height=20mm,scale=0.5]{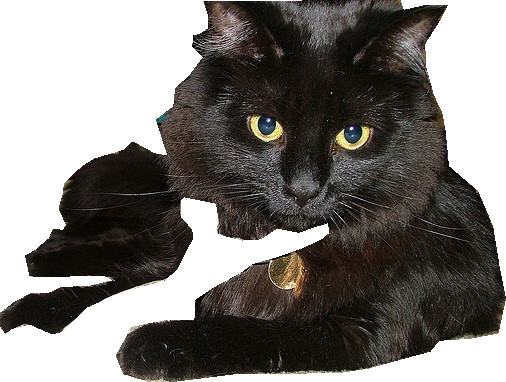}
	\end{subfigure}%
	\begin{subfigure}{.16\maxspace}
	    \centering 
		\includegraphics[height=20mm,scale=0.5]{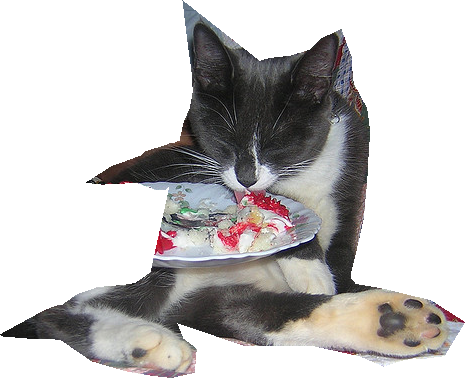}
	\end{subfigure}%
	\begin{subfigure}{.16\maxspace}
	    \centering
		\includegraphics[height=20mm,scale=0.5]{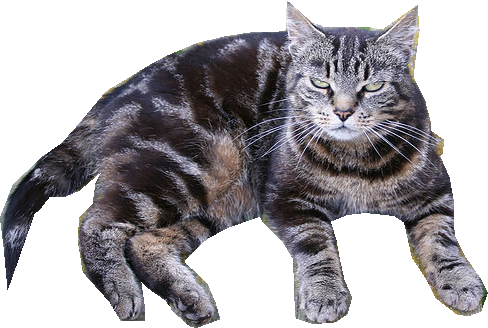}
	\end{subfigure}%
	\begin{subfigure}{.16\maxspace}
	    \centering
		\includegraphics[height=20mm,scale=0.5]{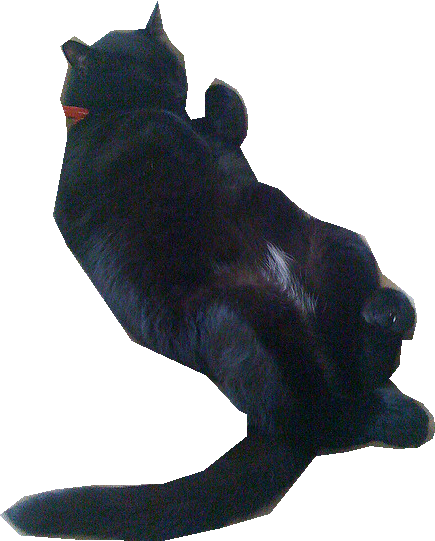}
	\end{subfigure}%
\par\medskip	
	\begin{subfigure}{.2\maxspace}
	    \centering
		\includegraphics[height=20mm,scale=0.2]{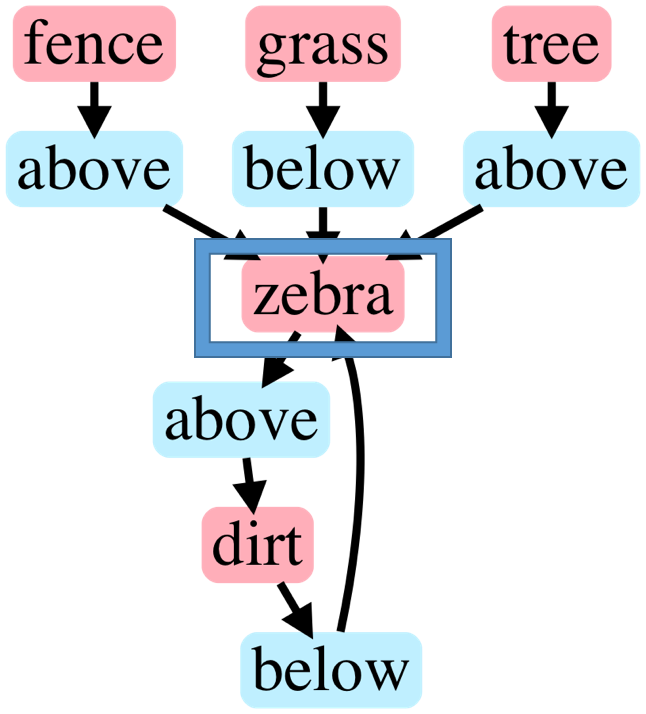} \\
	\end{subfigure}%
	\begin{subfigure}{.16\maxspace}
		\centering
		\includegraphics[height=20mm,scale=0.3]{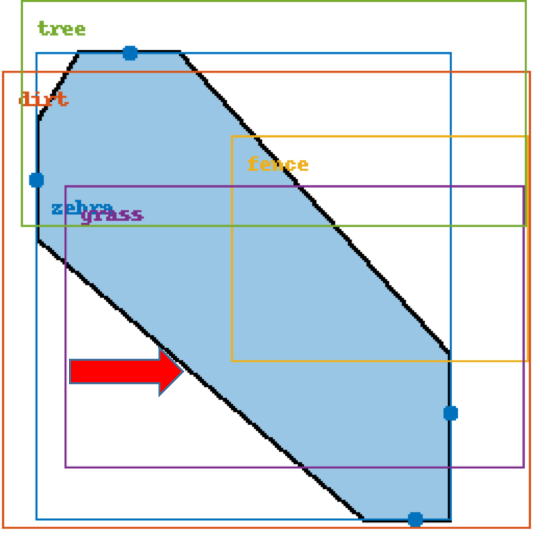} \\
	\end{subfigure}%
	\begin{subfigure}{.16\maxspace}
	    \centering
		\includegraphics[width=20mm,scale=1]{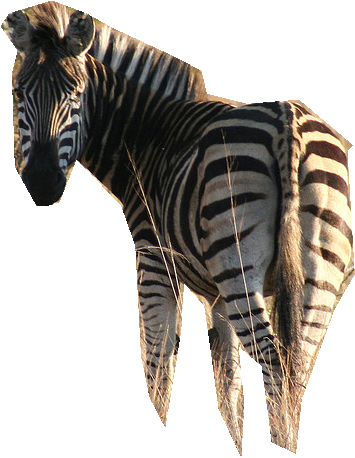} \\
	\end{subfigure}%
	\begin{subfigure}{.16\maxspace}
	    \centering 
		\includegraphics[height=20mm,scale=1]{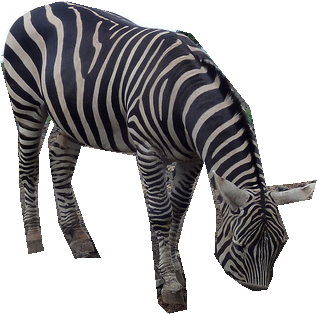} \\
	\end{subfigure}%
	\begin{subfigure}{.16\maxspace}
	    \centering
		\includegraphics[height=20mm,scale=1]{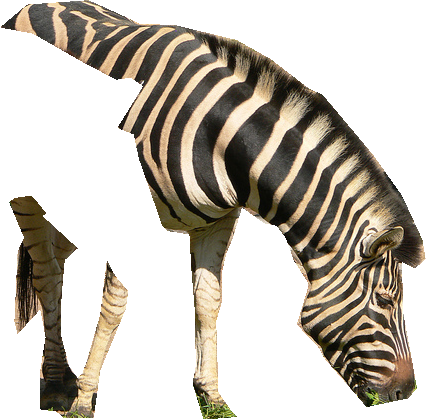} \\
	\end{subfigure}%
	\begin{subfigure}{.16\maxspace}
	    \centering
		\includegraphics[height=20mm,scale=1]{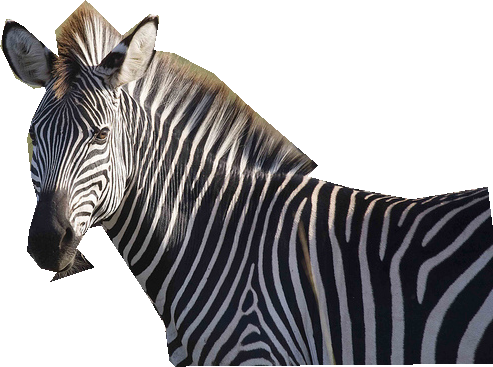} \\
	\end{subfigure}%
\par\medskip	
	\begin{subfigure}{.2\maxspace}
	    \centering
		\includegraphics[height=20mm,scale=0.2]{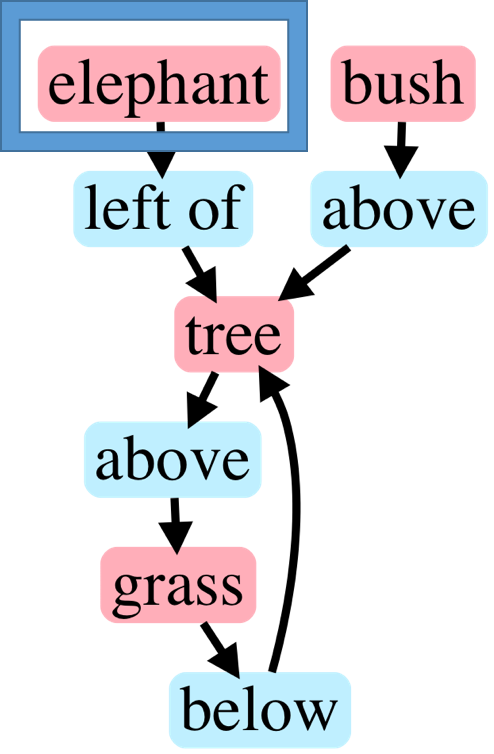} \\
	\end{subfigure}%
	\begin{subfigure}{.16\maxspace}
		\centering
		\includegraphics[height=20mm,scale=0.3]{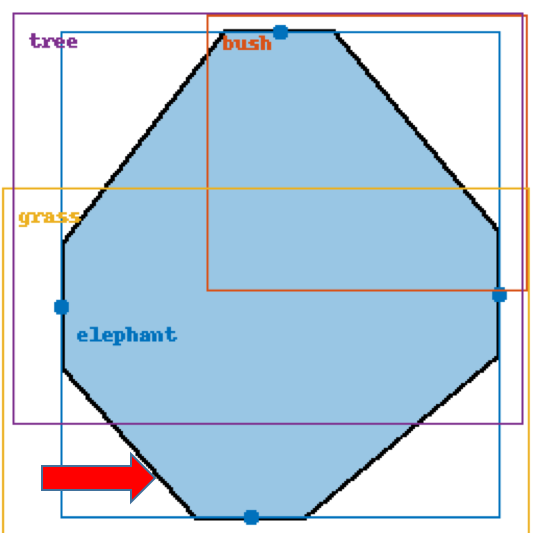} \\
	\end{subfigure}%
	\begin{subfigure}{.16\maxspace}
	    \centering
		\includegraphics[height=15mm,scale=1]{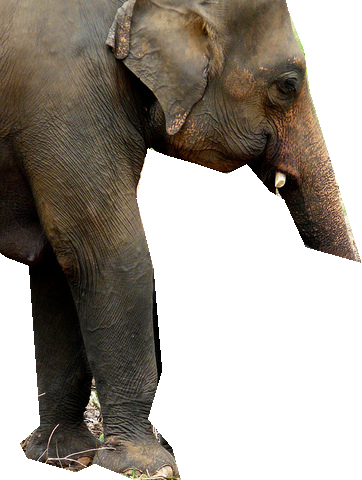} \\
	\end{subfigure}%
	\begin{subfigure}{.16\maxspace}
	    \centering 
		\includegraphics[width=20mm,scale=1]{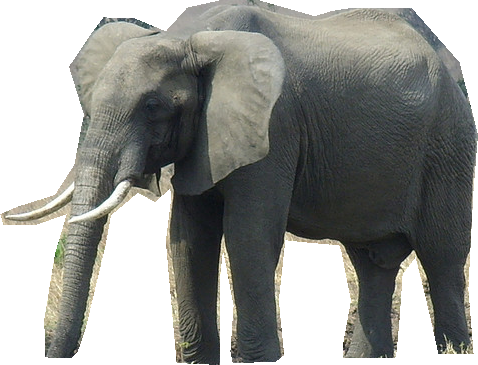} \\
	\end{subfigure}%
	\begin{subfigure}{.16\maxspace}
	    \centering
		\includegraphics[width=20mm,scale=1]{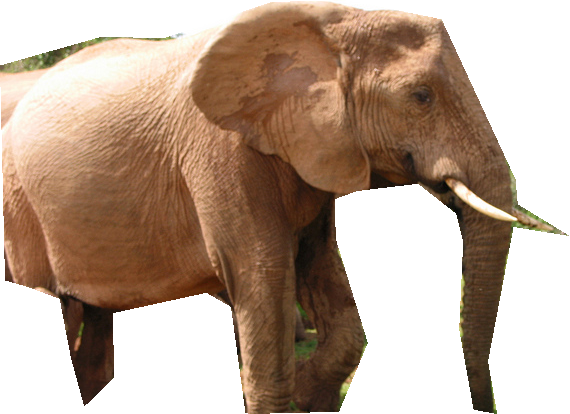} \\
	\end{subfigure}%
	\begin{subfigure}{.16\maxspace}
	    \centering
		\includegraphics[width=20mm,scale=1]{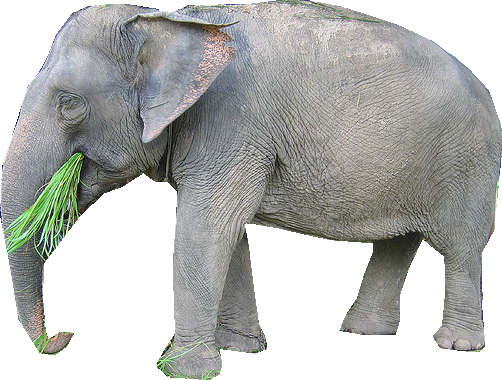} \\
	\end{subfigure}%
\par\medskip	
	\begin{subfigure}{.2\maxspace}
	    \centering
		\includegraphics[height=20mm,scale=0.2]{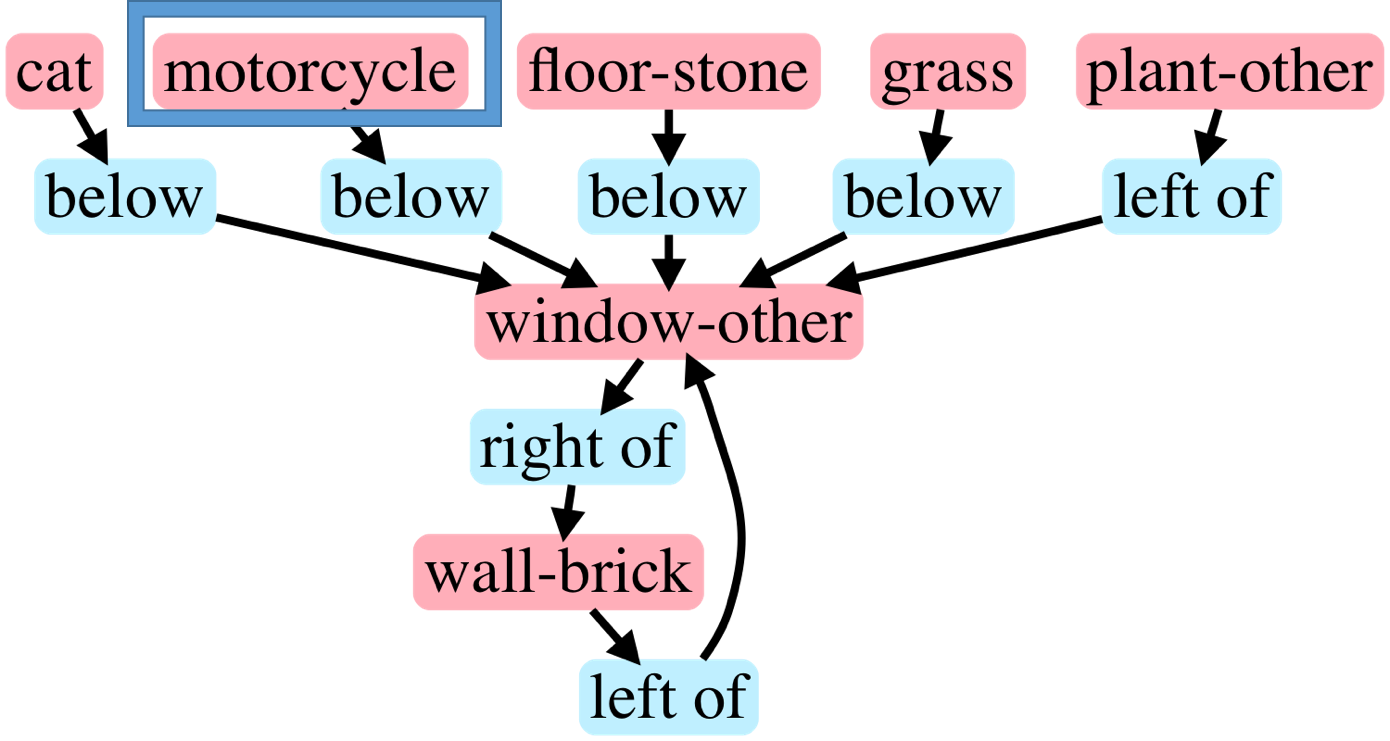} \\
	\end{subfigure}%
	\begin{subfigure}{.16\maxspace}
		\centering
		\includegraphics[height=20mm,scale=0.3]{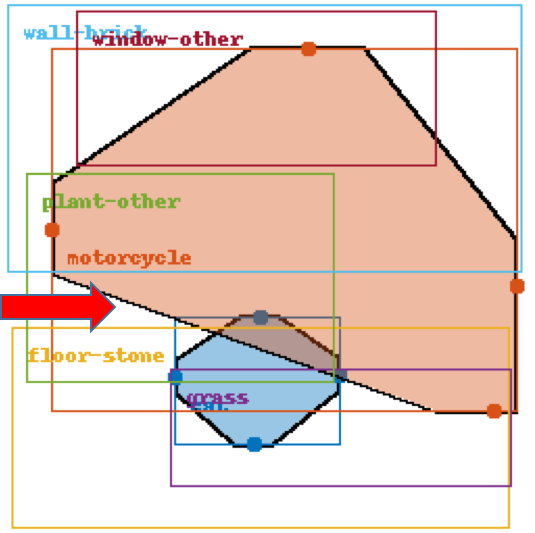} \\
	\end{subfigure}%
	\begin{subfigure}{.16\maxspace}
	    \centering
		\includegraphics[width=25mm,scale=1]{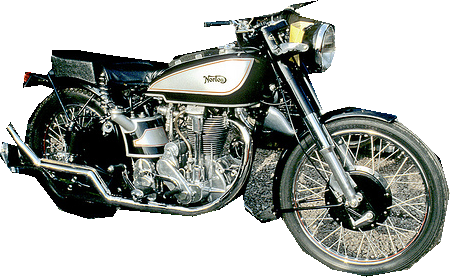} \\
	\end{subfigure}%
	\begin{subfigure}{.16\maxspace}
	    \centering 
		\includegraphics[width=20mm,scale=1]{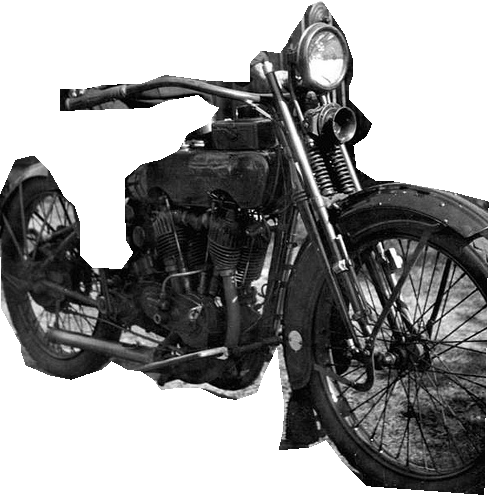} \\
	\end{subfigure}%
	\begin{subfigure}{.16\maxspace}
	    \centering
		\includegraphics[width=25mm,scale=1]{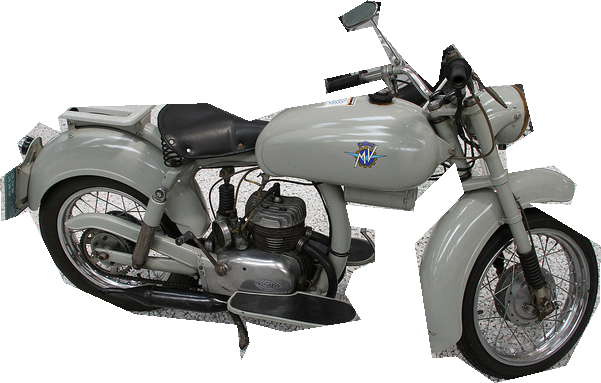} \\
	\end{subfigure}%
	\begin{subfigure}{.16\maxspace}
	    \centering
		\includegraphics[width=25mm,scale=1]{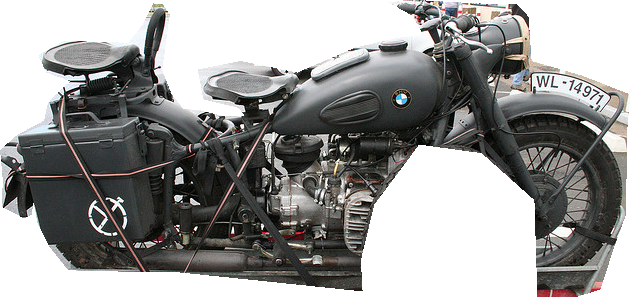} \\
	\end{subfigure}%
\par\medskip	
	\begin{subfigure}{.2\maxspace}
	    \centering
		\includegraphics[height=20mm,scale=0.2]{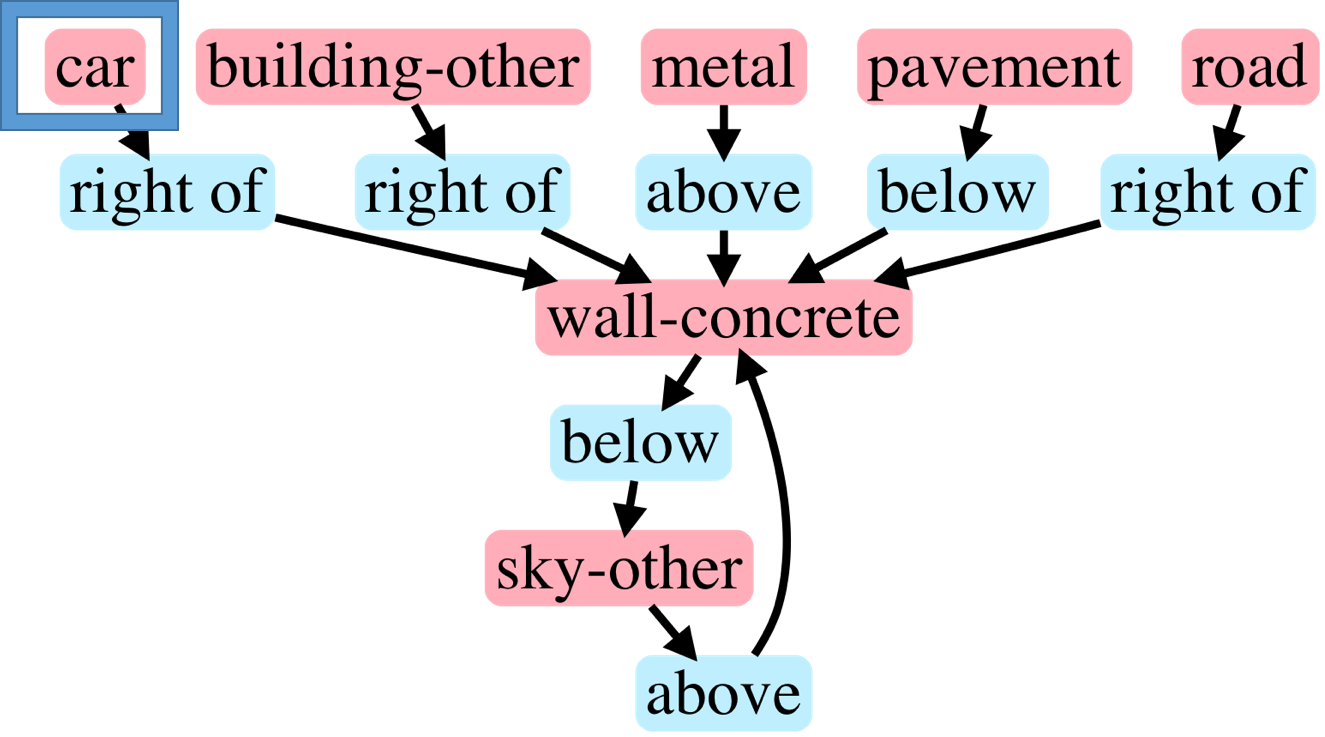} \\
	\end{subfigure}%
	\begin{subfigure}{.16\maxspace}
		\centering
		\includegraphics[height=20mm,scale=0.3]{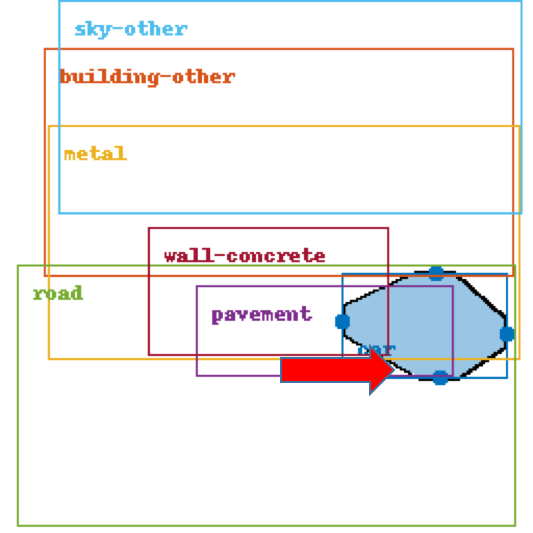} \\
	\end{subfigure}%
	\begin{subfigure}{.16\maxspace}
	    \centering
		\includegraphics[width=25mm,scale=1]{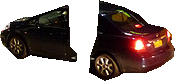} \\
	\end{subfigure}%
	\begin{subfigure}{.16\maxspace}
	    \centering 
		\includegraphics[width=25mm,scale=1]{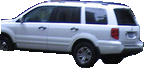} \\
	\end{subfigure}%
	\begin{subfigure}{.16\maxspace}
	    \centering
		\includegraphics[width=25mm,scale=1]{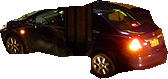} \\
	\end{subfigure}%
	\begin{subfigure}{.16\maxspace}
	    \centering
		\includegraphics[width=25mm,scale=1]{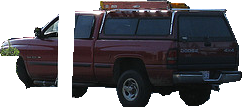} \\
	\end{subfigure}%
\par\medskip
	\begin{subfigure}{.2\maxspace}
	    \centering
		\includegraphics[height=20mm,scale=0.2]{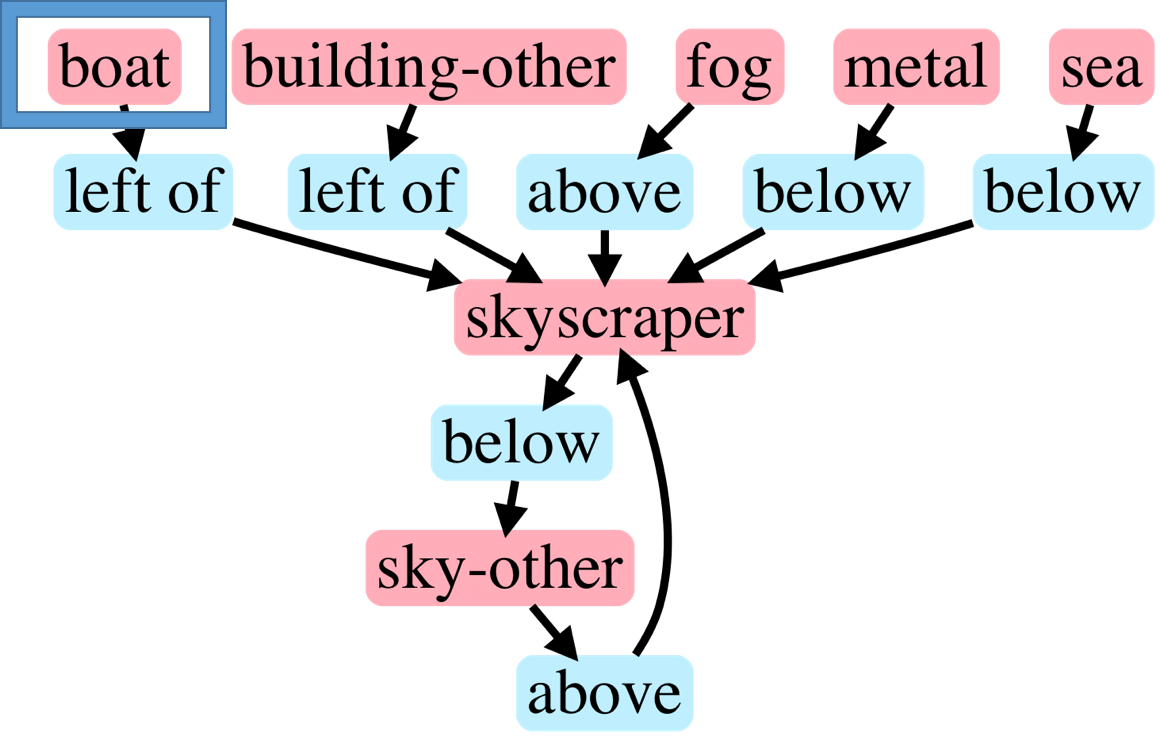} \\
	\end{subfigure}%
	\begin{subfigure}{.16\maxspace}
		\centering
		\includegraphics[height=20mm,scale=0.3]{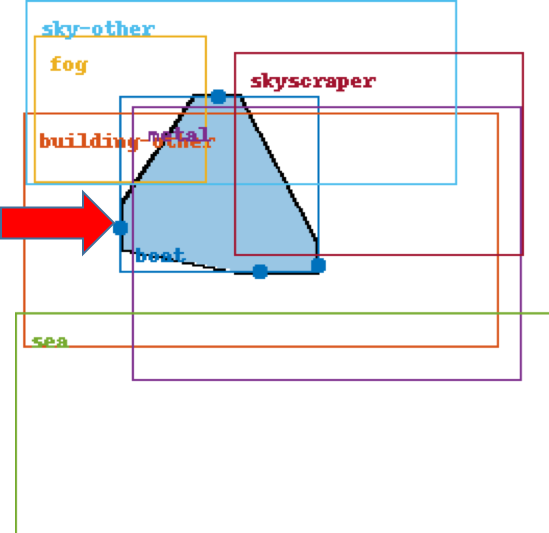} \\
	\end{subfigure}%
	\begin{subfigure}{.16\maxspace}
	    \centering
		\includegraphics[width=20mm,scale=1]{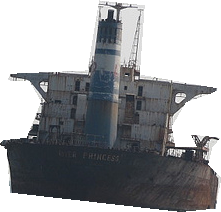} \\
	\end{subfigure}%
	\begin{subfigure}{.16\maxspace}
	    \centering 
		\includegraphics[width=25mm,scale=1]{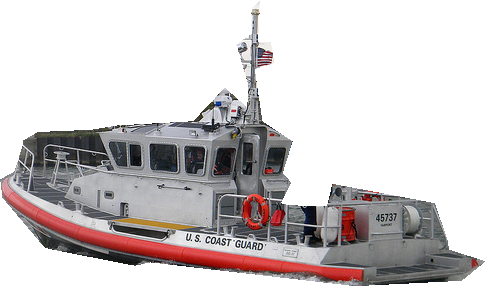} \\
	\end{subfigure}%
	\begin{subfigure}{.16\maxspace}
	    \centering
		\includegraphics[width=25mm,scale=1]{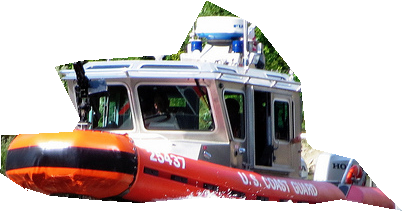} \\
	\end{subfigure}%
	\begin{subfigure}{.16\maxspace}
	    \centering
		\includegraphics[width=25mm,scale=1]{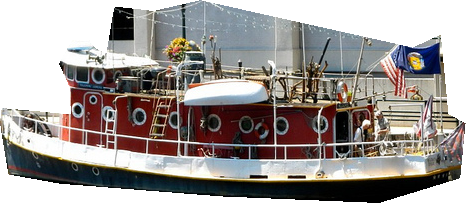} \\
	\end{subfigure}%
\par\medskip	
	\begin{subfigure}{.2\maxspace}
	    \centering
		\includegraphics[height=20mm,scale=0.2]{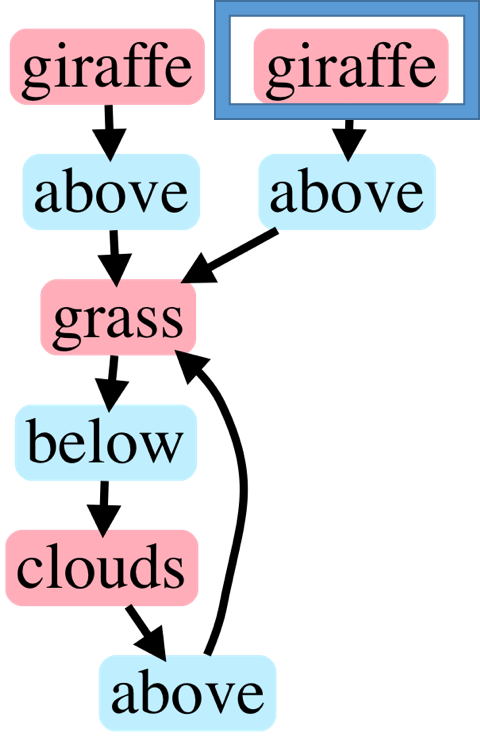} \\
		\par\smallskip
		\par\bigskip
		\caption{Scene Graph}
	\end{subfigure}%
	\begin{subfigure}{.16\maxspace}
		\centering
		\includegraphics[height=20mm,scale=0.3]{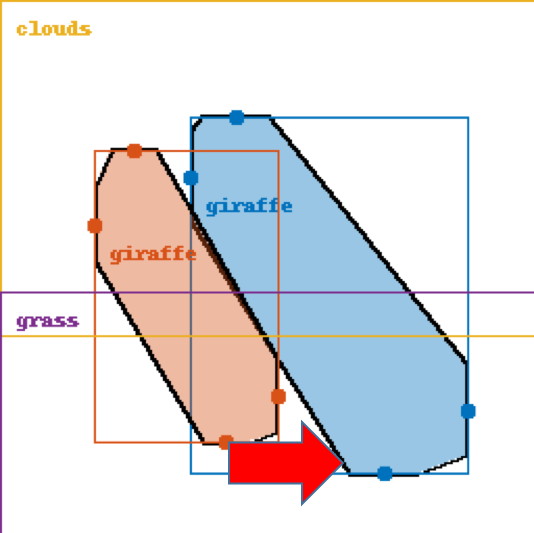} \\
		\par\smallskip
		\par\bigskip
		\caption{Predicted Octagons}
	\end{subfigure}%
	\begin{subfigure}{.16\maxspace}
	    \centering
		\includegraphics[height=20mm,scale=1]{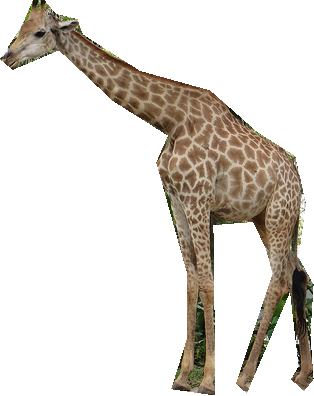} \\
		\par\bigskip
		\par\smallskip
		\caption{Rank 1 Retrieval}
	\end{subfigure}%
	\begin{subfigure}{.16\maxspace}
	    \centering 
		\includegraphics[height=20mm,scale=1]{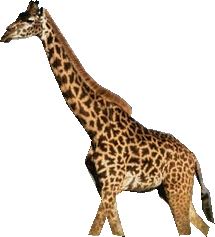} \\
		\par\smallskip
		\par\bigskip
		\caption{Rank 2 Retrieval}
	\end{subfigure}%
	\begin{subfigure}{.16\maxspace}
	    \centering
		\includegraphics[height=20mm,scale=1]{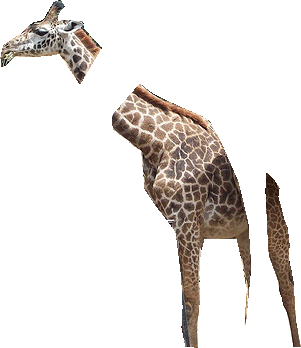}\\
		\par\smallskip
		\par\bigskip
		\caption{Rank 3 Retrieval}
	\end{subfigure}%
	\begin{subfigure}{.16\maxspace}
	    \centering
		\includegraphics[height=20mm,scale=1]{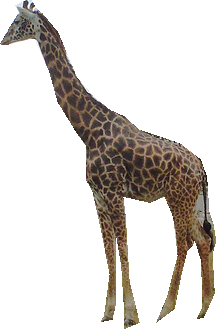} \\
		\par\smallskip
		\par\bigskip
		\caption{Rank 4 Retrieval} 
	\end{subfigure}%
	\vspace{-0.75\baselineskip}
	\par\medskip
	\centering\caption{Extreme Point based patch retrieval results. From left to right for each row - (a) Input scene graph, (b) Octagons generated from the predicted extreme points per object, (c)-(f) Top retrievals obtained using the proposed extreme point based method ranked in the order of preference (ie from best to worst). Note that the object being retrieved form the scene graph is highlighted using a blue rectangular box and the corresponding predicted octagon is indicated with a red arrow. Using the proposed method, the shape and pose of the top retrievals are very similar to the predicted octagons.}
	\label{fig:comparison}
\end{figure*}

For every predicted object in the validation set, we use the extreme point based retrieval method to extract the top 5 matches from the retrieval database. A few examples of such retrievals across different categories are shown in Figure \ref{fig:comparison}.  Qualitatively, note that the shape and pose of the retrieved patches closely match the predicted octagons. Additionally, we measured the IoU between the ground truth object mask and the retrieved patch mask (Table \ref{tab:retrieval}), which demonstrates that our extreme point-based retrieval is more accurate than other approaches. The Top-1 score is the IoU computed using the best retrieval while the Top-5 score is the highest IoU among the 5 retrievals. The scores are averaged across the 23 categories for all the objects in the validation set.

{
\section{Conclusion}
\label{sec:conclusion}
While scene graphs are compact representations of images, methods that can accurately use this representation for tasks such as image rendering or patch retrieval are not well developed. We've introduced several new contributions, including heuristic-based data augmentation, and extreme-point supervision, that demonstrate improved performance in utilizing scene graphs for generating realistic scene layouts. We additionally use the extreme point prediction module as a way to perform patch retrieval with only the scene graph as input. As graph-based tasks increase in number and diversity \cite{battaglia2018relational}, we expect our contributions to generalize to other graph-conditioned algorithms. Summarily, we hope to build upon our current results by leveraging recent progress in generating high-resolution photo-realistic images \cite{karras2017progressive, brock2018large}.
\clearpage
}
{\balance
\small
\bibliographystyle{ieee}
\bibliography{main}
}

\end{document}